\documentclass[10pt,journal,compsoc,twoside]{IEEEtran}
\usepackage[linesnumbered,ruled,vlined]{algorithm2e}

\SetCommentSty{mycommfont}
\usepackage{xcolor}
\usepackage{amssymb}
\usepackage{multirow}
\usepackage{mathtools}

\definecolor{orange}{rgb}{0.6, 0.5, 0.0}
\definecolor{cfgreen}{rgb}{0.0, 0.42, 0.24}
\definecolor{ppocra}{rgb}{1,0.45, 0.00}

\ifCLASSOPTIONcompsoc
  \usepackage[nocompress]{cite}
\else
  \usepackage{cite}
\fi
\ifCLASSINFOpdf
  \usepackage[pdftex]{graphicx}
\else
\fi
\usepackage{amsmath}
\usepackage{array}
\def\etal{\emph{et al.}}
\def\eg{\emph{e.g.}}
\def\ie{\emph{i.e.}}
\def\etc{\emph{etc.}}
\newcommand{\hide}[1]{}

\begin{document}

\title{A Sparse and Locally Coherent Morphable Face Model for Dense Semantic Correspondence Across Heterogeneous 3D Faces}


\author{Claudio Ferrari,~\IEEEmembership{Member,~IEEE,}
        Stefano Berretti,~\IEEEmembership{Senior,~IEEE,}
        Pietro Pala,~\IEEEmembership{Senior,~IEEE,} \\
        and~Alberto Del~Bimbo,~\IEEEmembership{Senior,~IEEE}
\IEEEcompsocitemizethanks{\IEEEcompsocthanksitem C. Ferrari, S. Berretti, P. Pala and A. Del~Bimbo are with the Department of Information Engineering, University of Florence, Florence, Italy, 50139.\protect\\
E-mail: claudio.ferrari@unifi.it}
\thanks{Manuscript received ; revised.}}

\markboth{IEEE Transactions on ?,~Vol.~??, No.~??, November~2020}%
{C. Ferrari \MakeLowercase{\etal}: Semantic Annotation Transfer for Dense 3D Face Correspondence}

\IEEEtitleabstractindextext{%
\begin{abstract}
The 3D Morphable Model (3DMM) is a powerful statistical tool for representing 3D face shapes. To build a 3DMM, a training set of face scans in full point-to-point correspondence is required, and its modeling capabilities directly depend on the variability contained in the training data. Thus, to increase the descriptive power of the 3DMM, establishing a dense correspondence across heterogeneous scans with sufficient diversity in terms of identities, ethnicities, or expressions becomes essential. In this manuscript, we present a fully automatic approach that leverages a 3DMM to transfer its dense semantic annotation across raw 3D faces, establishing a dense correspondence between them. We propose a novel formulation to learn a set of sparse deformation components with local support on the face that, together with an original non-rigid deformation algorithm, allow the 3DMM to precisely fit unseen faces and transfer its semantic annotation. We extensively experimented our approach, showing it can effectively generalize to highly diverse samples and accurately establish a dense correspondence even in presence of complex facial expressions. The accuracy of the dense registration is demonstrated by building a heterogeneous, large-scale 3DMM from more than 9,000 fully registered scans obtained by joining three large datasets together.
\end{abstract}

\begin{IEEEkeywords}
3D Morphable Model, Sparse Components Learning, Dense Correspondence
\end{IEEEkeywords}}

\maketitle

\IEEEdisplaynontitleabstractindextext
\IEEEpeerreviewmaketitle

\IEEEraisesectionheading{\section{Introduction}\label{sec:introduction}}
\IEEEPARstart{I}{n} face analysis, one widely used statistical model that gained increasing interest throughout the years is the 3D Morphable Model (3DMM)~\cite{Blanz:1999}. Thanks to its ability to represent and manipulate 3D face shapes, it found successful application in a variety of tasks, spanning from single view reconstruction~\cite{tun2018extreme, galteri2019deep, gecer2019ganfit, Tu-TMM:2020, Tu2019Joint3F, galteri2019coarse}, computer graphics~\cite{neumann:2013, Zhao-NIPS2017, ferrari-2018}, to biometrics~\cite{berretti:2010, ferrari:2016, masi:2014, zhao20183d, ramanathan:2006, Zhao-ijcai2018, zhao2018towards, zhao2020recognizing, tu2021joint} or medical imaging~\cite{amberg:2008, blanz:2003, Cao-cvpr:2018, ferrari2021inner}. In general, a 3DMM can be constructed by computing an average face and a set of \textit{deformation components}, learned from an ensemble of fully registered face scans. Its capability of generating variegated, novel faces strongly depends on the variability of the training data. So, being able to collect heterogeneous and abundant data appears decisive to build a powerful model~\cite{Booth:2017a}. Data heterogeneity can be obtained by including numerous identities of different gender, ethnicity, or age, preferably including expressions. However, this first requires putting the raw scans in a dense, point-to-point correspondence. Given the burdensome process of 3D data acquisition, a possible workaround is that of combining scans coming from different, existing, 3D face datasets. Unfortunately, the very different surface characteristics as induced by scanning devices further complicate the dense registration problem.

Among other applications, 3DMMs have been used as statistical priors to establish such correspondence among raw 3D faces. This is achieved by first fitting a \textit{template} face to the target scans using the deformation components, and then transferring the template's topology to them. This operation aims at mitigating the irregularities of raw scans, in terms of number and disposition of acquired points, by associating each point of the template with its most \textit{similar} one in the target. The similarity can account for different criteria, like a point-to-point distance, or the similarity between surface attributes (\eg,~normal, color, semantic annotations,~\etc). Then, the topology and semantic annotation of the template (\eg,~facial landmarks) can be transferred to the targets as well. Repeating this operation on a set of scans by using the same template induces, transitively, a dense correspondence among all of them. Nonetheless, standard 3DMMs are usually built from small training sets, \ie,~including few identities or neglecting expressions, hence limiting their generalization capability. This is due both to the difficulty of collecting 3D data as mentioned above, and the lack of effective non-rigid point cloud registration methods that can handle large topological changes, as those occurring in human faces. From these considerations, it is evident that putting a large set of 3D faces in dense correspondence to increase the data diversity requires the capability of adapting to novel face shapes with heterogeneous characteristics, and identities that are not part of the training data. Unfortunately, even the most recent non-linear 3DMMs, as those based on convolutional mesh operators~\cite{ranjan2018generating, bouritsas2019neural}, do not provide a robust answer yet, though they proved extremely effective in interpolating between samples belonging to the training data.

To tackle the dense registration problem, we observe it is essential to elaborate appropriate solutions to: (\emph{i}) learn deformation components that capture as much variability in the training data as possible. In addition, the capability of generalizing to diverse samples despite the lack of variegated training data is also a non-negligible requirement; (\emph{ii}) deform the template by using the learned components to fit a target scan. Assuming the template might largely deviate from the targets, a sufficiently general solution to match the surfaces is of utmost importance to maintain a consistent topology across the scans.
It has been shown that both for (\emph{i}) and (\emph{ii}), spatial locality is a desirable property as opposite to 3DMMs that use a global support of the face. 
This increases the model flexibility by decorrelating complex movements into local, separate deformations~\cite{Luthi-tpami:2018}, while also improving the robustness to expressions and noise~\cite{brunton2014multilinear}.

In this paper, we provide original solutions to the above points (\emph{i})-(\emph{ii}). 
First, we propose a new method for learning localized yet anatomically meaningful deformation components, named \textit{Sparse and Locally Coherent} (SLC), that can widely generalize to unseen shapes. We formulate the components learning by taking a completely new look at the training data, treating each vertex in the 3D faces as an independent sample. This change in the learning paradigm allows us to model the variability of each vertex coordinate independently from the others, reducing the impact of the correlation between different regions of the same face, and the dependency from the number and variability of training samples. As a consequence of disentangling complex facial movements, the learned components result sparse and spatially localized, significantly enlarging the spectrum of possible deformations.

Then, we design a solution that uses the learned components to deform a template to target 3D faces by means of an iterative optimization, named \textit{Non-Rigid Fitting} (NRF). Its peculiarity consists in a novel strategy to establish a preliminary point-to-point correspondence between template and target, which assigns each vertex of the template to the centroid of its $k$-nearest neighbors in the target (Voronoi region). Together with a dynamic outliers rejection policy, it allows the template to smoothly adapt to large shape differences. This represents the first example of a method capable of handling both neutral and strongly expressive faces, as well as diversity in terms of resolution, mesh topology and noise, without requiring landmarks to initialize the deformation. Finally, we transfer the topology of the deformed model to the target faces, putting them in dense correspondence.

In the experiments, we first provide a comparison between the proposed SLC and state-of-the-art approaches in the task of 3D face modeling and reconstruction. Then, we evaluate the proposed methods for components learning and registration against state-of-the-art approaches on three large and heterogeneous datasets. Finally, we show the registered scans of the three datasets are accurate enough to derive a \emph{Heterogeneous Large-Scale} 3DMM with improved modeling capabilities.
Overall, the main contributions are:
\begin{itemize}
\item We propose a novel 3DMM learning solution (SLC) in which each vertex across the training scans represents an independent sample. This allows us to learn sparse and localized deformations, leading to a model with remarkable generalization capability to unseen shapes. This contributes a clear discontinuity with respect to previous works that instead consider the overall face support as training sample.
\item We design an algorithm to accurately fit a 3DMM to raw 3D faces (NRF) that can handle large topological changes as induced by expressions without requiring a landmark-based initialization.
\item We apply the methods above to establish a dense correspondence over more than $9,000$ raw scans of $721$ subjects from three different datasets, overcoming critical drawbacks associated with data collection and handling of expressive scans.
\end{itemize}

The rest of the paper is organized as follows: in Section~\ref{sec:related-work}, we summarize the works in the literature that are more close to our proposed solution. In Section~\ref{sec:3DMM}, we formulate the new approach for learning localized deformation components of a 3DMM. Then, description of the 3DMM inclusion in a framework for dense semantic annotation transfer is addressed in Section~\ref{sec:dense_transfer}; Extended experimental results are reported in Section~\ref{sec:experimental-results}. Finally, discussion and conclusions are sketched in Section~\ref{sec:conclusion}. Code and data are publicly available at  https://github.com/clferrari/SLC-3DMM.

\section{Related Work}\label{sec:related-work}
We summarize relevant works in the literature that establish dense correspondence across 3D face scans, construct statistical 3D face models as well as large scale 3DMMs. For a comprehensive survey on 3DMMs, we refer to~\cite{egger20203dmmsurvey}.

\textbf{Constructing Statistical 3D Face Models} -- Two aspects have a major relevance in characterizing the different methods for 3DMM construction: (\emph{i}) The human face variability captured by the training scans, which directly depends on their number and heterogeneity; (\emph{ii}) The approach employed to learn the components.
The former point is relevant as the standard 3DMM cannot replicate deformations that do not appear in the training data, like facial expressions~\cite{ferrari:2015}. The second point determines to what extent the 3DMM is able to capture the latent structure contained in the data and generalize to unseen samples.
Most of the literature on 3DMM construction relies on the seminal work of Blanz and Vetter~\cite{Blanz:1999}. In their solution, the shape and texture of a training set of 3D faces were transformed into a vector space representation based on PCA, each face representing a training sample. This 3DMM proposal was further refined into the Basel Face Model (BFM) by Paysan~\etal~\cite{paysan:2009}.
Cao~\etal~\cite{Cao-facewarehouse:2014} proposed a popular multi-linear 3D face model, called FaceWarehouse, that introduced expressive scans in the 3DMM training set.
More recently, Li~\etal~\cite{FLAME:SiggraphAsia2017} proposed FLAME, a powerful multi-linear PCA model composed of shape, expression blendshapes and pose parameters that are learned separately from 4D sequences. The latter improves upon the FaceWarehouse model by using a larger number of training scans, demonstrating the advantage of abundance of data.
Some later methods investigated different ways to learn the deformation components.  Brunton~\etal~\cite{brunton2014multilinear} defined a multi-linear model based on wavelet decomposition, and showed the advantage of learning localized and decorrelated components to deal with identity and expression variations in a 3D-3D fitting scenario. Still, to fit expressive faces, the deformation needed to be initialized with a set of pre-detected landmarks. 
The work by L{\"{u}}thi~\etal~\cite{Luthi-tpami:2018} is also capable of modeling local and spatially uncorrelated deformations by using a Gaussian Process 3DMM to generalize the PCA-based shape model. They also elaborated on the importance of de-correlating facial movements to achieve more flexibility, but the learned local deformations did not fully respect the anatomical structure of the face.
Following similar ideas, Neumann~\etal~\cite{neumann:2013} proposed a sparse variant of PCA with additional local support constraints to achieve localized yet realistic deformations. However, such deformations were learned on mesh sequences of single subjects, and used mainly for artistic and animation purposes.
The Dictionary Learning-based 3DMM (DL-3DMM) presented in~\cite{ferrari:2015}, instead, generalized the PCA model by learning a dictionary of deformation components, joining shape and expression variations into a single model. The DL-3DMM increased the modeling capabilities by removing the orthogonality constraint as imposed by PCA. Despite using a similar learning formulation with respect to the proposed one, each 3D face still represents a training sample, which gets reflected in limited generalization in case of small training sets.

Recently, several solutions have been presented that apply Convolutional Neural Networks to learn non-linear 3D face models. These methods can regress shape and texture parameters directly from an input photo~\cite{Sengupta-sfsnet:2018}, or UV maps~\cite{Bagautdinov-vae:2018}. Some solutions added the capability of modeling extreme expressions using convolutional mesh autoencoders~\cite{ranjan2018generating, Jiang-cvpr:2019}.
Ranjan~\etal~\cite{ranjan2018generating} proposed to learn a non-linear face representation using spectral convolutions on a mesh surface, and introducing mesh sampling operations that enable a hierarchical mesh representation in an encoder-decoder architecture. The resulting model captures non-linear variations in shape and expression at multiple scales.
This work was extended by Jiang~\etal~\cite{Jiang-cvpr:2019} to decompose a 3D face into identity and expression parts.
Bouritsas~\etal~\cite{bouritsas2019neural} introduced the \textit{spiral convolution} operator that acts directly on the 3D mesh and explicitly models the bias of the fixed underlying graph. The spiral operator enforces consistent local orderings of the graph vertices, thus breaking the permutation invariance property that is adopted by prior works on Graph Neural Networks. However, these solutions still lack sufficient generalization, and face several limitations when unseen data are considered.
Liu~\etal~\cite{Liu-iccv:2019} learned a non-linear face model from a large set of raw 3D scans. They used the PointNet architecture~\cite{Charles-pointnet:2017} to convert point clouds to identity and expression latent representations, while establishing a dense point-to-point correspondence among them. Incidentally, this is the unique method in the literature that deals with scans from different databases. The training data also included synthetic scans for which a dense correspondence is known, so resembling a semi-supervised setup. Further, employing the PointNet architecture resulted in a global shape model.

With respect to the above solutions, the novelty of our solution is that the variation of each vertex coordinate across the training scans is treated as an independent sample, which brings important advantages. The dependency from the training scans is significantly reduced, and the resulting model can learn a wide and variegated range of deformations even from small training sets. Moreover, by de-correlating motions occurring in different regions of the same face, we can learn sparse yet localized deformations and overcome the generalization problem to a great extent.

\textbf{Dense Correspondence between 3D Face Scans} -- Determining a dense correspondence between 3D point clouds can be seen as a particular case of a rigid/non-rigid registration problem.
State-of-the-art methods can be categorized according to the fact they:
(\emph{i}) compute surface descriptors that can be used to define few landmarks of the face~\cite{Creusot-ijcv:2013, Lu-fg:2006, Segundo-tcyb:2010, Perakis-tpami:2013, Perakis-pr:2014}, perform region matching~\cite{Fan-cvpr:2019, salazar2014fully, Sun-iccv:2001} or initialize a graph-matching procedure~\cite{Zeng-registration:2016, Zeng-registration:2010};
(\emph{ii}) resolve for an objective function to find a mapping between fiducial points~\cite{Amberg-nicp:2007, booth:2016, Li-correspondence:2008}, or
(\emph{iii}) use a 3D model of the face to transfer a dense semantic labeling.
In the following, we focus on the latter category. An in depth discussion on shape correspondence methods can be found in~\cite{Maiseli-survey:2017, Tam-survey:2013}.

Several solutions used a 3D model of the face to transfer a set of reference labels to a set of unlabeled scans. For example, Paysan~\etal~\cite{paysan:2009} proposed a registration method based on the Non-rigid Iterative Closest Point (NICP)~\cite{Amberg-nicp:2007} algorithm. Since NICP cannot handle large missing regions and topological variations, facial expressions were not accounted for in this work. Gilani~\etal~\cite{Gilani-correspondence:2015} proposed to evolve level set curves with adaptive geometric speed functions to put in correspondence a large number of landmarks. The method was refined in~\cite{Gilani-correspondence:2017}, where a multi-linear algorithm was used to establish a dense point-to-point correspondence over a population of 3D faces varying in identity and facial expression. To deal with expressions, a set of facial landmarks were detected using a deep network and employed to initialize the deformation process. Gilani~\etal~\cite{Gilani-dense:2018} proposed an algorithm based on the subsequent refinement of automatically detected sparse correspondences on the outer boundary of 3D faces. However, only neutral and few mild expressive scans were put in dense correspondence.
Fan~\etal~\cite{Fan-eccv:2018} proposed a template warping strategy to achieve semantic and topological correspondence based on the Go-ICP~\cite{Yang-goICP:2016} algorithm. 
This approach does not rely on a sparse set of landmarks for initialization, but it cannot deal with facial expressions.
Fan~\etal~\cite{Fan-cvpr:2019} achieved dense correspondence by deforming a template face with locally rigid motions, that are guided by sparse seed points. To account for facial expressions, the deformation is initialized using four landmarks on the mouth region that are manually labeled in the target scans. Salazar~\etal~\cite{salazar2014fully} also used a set of pre-detected landmarks to establish point-to-point correspondence across expressive faces.

All the works above dealing with expressive shapes require a landmark based initialization. Differently, our solution can put in dense correspondence neutral as well as strongly expressive scans without using any landmarks. This is a valuable property as accurately detecting landmarks on 3D faces is by itself a challenging problem. In addition, our strategy allows us to put in correspondence and join scans collected from different datasets. Differently, none of the above solutions addressed this issue.

\begin{figure*}[!t]
	\centering
	\includegraphics[width=0.93\linewidth]{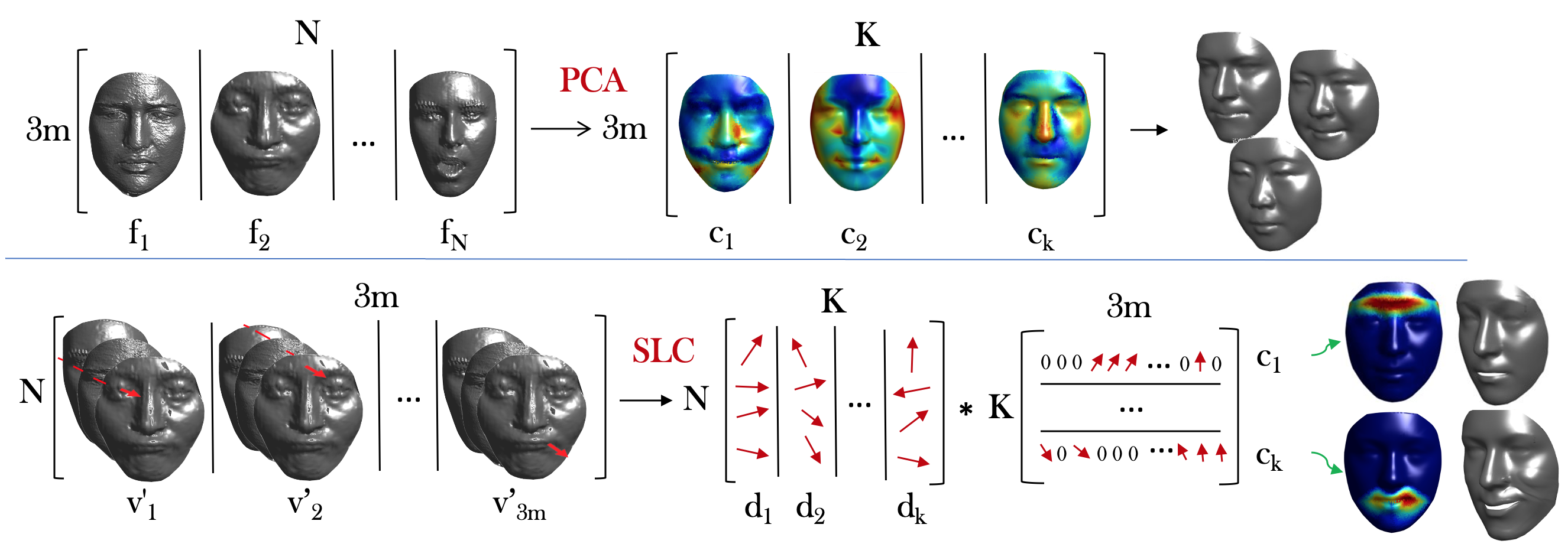}
	\caption{\label{fig:defField} Previous solutions, \eg, PCA~\cite{paysan20093d} and others, interpret each 3D face as a training sample, and learn a set of $k < N$ components (top row). In our solution (bottom row), we analyze each of the $3m$ coordinate displacements $\mathbf{v}'_i$ independently. A set of $k \ll 3m$ primary directions $\mathbf{d}_i$ is extracted from $\mathbf{V}'$. The coefficients $\mathbf{c}_i$ identify a way to expand the $k$ learned directions back to the $3m$ coordinates and deform the 3D model.}
\end{figure*}

\textbf{Large Scale Morphable Models} The first 3DMM constructed on a large set of training examples was presented by Booth~\etal~\cite{booth:2016, Booth:2017a}. They constructed the Large-Scale Facial Model (LSFM) from 9,663 distinct facial identities (about 12,000 scans in total), including variations in age, gender and ethnicity. All the scans were acquired in neutral expression from the same device. The dense registration was achieved using NICP by relying on a set of labeled landmarks.
Ploumpis~\etal~\cite{ploumpis2019combining} presented a general approach to combine 3DMMs from different parts of an object class into a single 3DMM. They fused the variability and facial details of the LSFM with the full head modelling of the Liverpool-York Head Model (LYHM)~\cite{dai-craniofacial:2017}, thus creating the Combined Face \& Head Model (CFHM). This provided a solution to the problem of combining existing models built using different templates that can only partly overlap, have different representation capabilities, and are constructed from different datasets. Finally, the work of Ploumpis~\etal~\cite{ploumpis2020towards} presented a complete model of the human head extending the previous ones with eyeballs and ears.

The models proposed in the works above have remarkable and promising modeling capabilities, demonstrating the advantage of learning a 3DMM from a large and variegated set of scans. On the other hand, collecting and processing the data required a huge effort. In addition, handling expressive scans still represents a difficulty that existing large scale models do not deal with, which evidently limits their potential. We argue our proposed work can be of great help in bridging this gap by providing a solution that allows combining scans with a large spectrum of variabilities, including expressions, coming from datasets that have been already collected.

\section{Learning Sparse and Locally Coherent Deformation Components}\label{sec:3DMM}
The general idea of 3DMMs is that of generating novel 3D faces $\mathbf{s}$ by deforming an average model $\mathbf{m}$ through a combination of the deformation components $\mathbf{c}_i$:
\begin{equation}
\label{eq:3dmmL}
\mathbf{s} = \mathbf{m} + \sum_{i=1}^{k} \mathbf{c}_i \alpha_i \; .
\end{equation}

\noindent
Although several alternatives to PCA have been proposed to learn the components $\mathbf{c}_i$, the founding concept is that they should capture relevant relationships in the data, representing specific face characteristics or attributes. These can either be related to structures that define the \emph{identity} characteristics, \eg, nose shape, femininity/masculinity, \etc. Otherwise, they can involve deformations of the movable face parts, \eg, mouth opening/closing, muscular movements, \etc, which are independent from the identity, and are associated to facial \emph{expressions}. In both the cases, the capability of generalizing to unseen deformations depends on both the information carried by training faces, and the approach employed to learn the components $\mathbf{c}_i$.

In this work, we aim at deriving a set of deformation components that (\emph{i}) model both identity and expression variations as defined above, (\emph{ii}) are spatially localized and, (\emph{iii}) can generalize to unseen shapes and deformations.
For this purpose, we developed a method to learn localized yet meaningful deformations based on a sparse decomposition of the training matrix.

\subsection{Problem Formulation}\label{sec:NMF}
Throughout this section, we assume the 3D faces are in full correspondence. We represent the geometry of a generic 3D face as a vector $\mathbf{f}_i = [x_1,y_1,z_1, \dots, x_m,y_m,z_m]^T \in \mathbb{R}^{3m}$ that contains the linearized $(x,y,z)$ coordinates of the $m$ vertices. Let $\mathbf{F} = \left[ \mathbf{f}_1 | , \dots, | \mathbf{f}_N \right] \in \mathbb{R}^{3m \times N}$ be the matrix of the $N$ training scans, each with $m$ vertices arranged column-wise. Our goal is to extract from $\mathbf{F}$ a set of sparse deformation components.
To this aim, we first compute the difference between training scans and the average 3D face:
\begin{align}
\mathbf{m} = \frac{1}{N} \sum_{i=1}^{N} \mathbf{f}_i \; , && \mathbf{v}_i = \mathbf{f}_i - \mathbf{m} \; , \enspace \forall \; \mathbf{f}_i \in \mathbf{F} \; .
\end{align}

\noindent
Each $\mathbf{v}_i$ represents the \textit{displacement field} that transforms the average model $\mathbf{m}$ into a training model $\mathbf{f}_i$. Such $\mathbf{v}_i$ are then stacked to form a new matrix $\mathbf{V} = \left[ \mathbf{v}_1 |, \dots, | \mathbf{v}_N \right] \in \mathbb{R}^{3m \times N}$.

In the literature, approaches that were proposed to build a MM, such as~\cite{Blanz:1999, koppen2018gaussian, ferrari:2015, Booth:2017, Tran-cvpr:2018, brunton:2014a}, all interpret each 3D face as a training sample, \ie, a point lying on a $3m$-dimensional manifold. Operating this way, the components estimation is heavily influenced by the number of available scans. In addition, correlations between different parts of the same face are captured while learning the components.

We are interested in learning a set of sparse deformation components in which the interactions between different face parts can be decorrelated.
Our solution builds upon the observation that the musculoskeletal structure induces neighboring vertices to move according to consistent patterns (principle of local consistency of motion), making it possible to approximate the movement of a local region with single motion vectors that we refer to as \textit{primary directions}. We leverage this property by first learning a corpus of primary deformation directions from the aligned training scans. Then, we learn how to expand each primary direction to a localized set of vertices. The local consistency of motion ensures each direction is expanded and applied to neighboring points, allowing us to model sparse and localized deformations yet respecting the facial anatomy. The process is illustrated in Figure~\ref{fig:defField}.
To this end, we change the way we look at the training data. Instead of using each scan $\mathbf{v}_i$ as a separate training sample, we analyze the displacements of each vertex coordinate across the $N$ scans independently.
Therefore, each sample becomes an $N$-dimensional point $\mathbf{v}'_i \in \mathbb{R}^N$.

\subsection{Building the SLC-3DMM}\label{subsec:build3dmm}
Let $\mathbf{V}' = \mathbf{V}^T \in \mathbb{R}^{N \times 3m}$ be the transposed training matrix, so that each sample is $\mathbf{v}'_{i} \in \mathbb{R}^{N}$. We wish to find a set of $k$ ($k \ll 3m$) primary directions $\mathbf{D} \in \mathbb{R}^{N \times k}$ and sparse expansion coefficients $\mathbf{C} = \left[ \mathbf{c}_1 |, \dots, | \mathbf{c}_{3m} \right] \in \mathbb{R}^{k \times 3m}$ that allow optimal reconstruction of the input data, \ie, such that $\left\| \mathbf{V}' - \mathbf{D} \mathbf{C} \right\|_2^2$ is minimized and $\mathbf{C}$ is sparse. The learned coefficients are then used to apply localized deformations.

To formalize the problem, we first need to elaborate on some prerequisites. Since the coefficients $\mathbf{C}$ will be used as deformation components, it is important to ensure they result in smooth surface deformations, \ie,~without sharp transitions to zero-valued elements. Moreover, we observe that each $\mathbf{v}'_i$ can include both positive or negative directions, which might occur concurrently in different facial regions: for example, in the ``Angry'' expression, upper lip raising and eyebrows lowering are observed simultaneously. To effectively de-correlate the motions, we want to avoid them to be jointly learned. Given the above, we formulate the primary directions and sparse expansion coefficients learning as a constrained Elastic-net regression:
\begin{gather}
\label{eq:DLSC}
\min_{\mathbf{c}_i , \mathbf{D}} \frac{1}{3m} \sum_{i=1}^{3m} \left (
\left\| \mathbf{v}'_i - \mathbf{D} \mathbf{c}_i \right\|_2^2 + \lambda_1 \left\|\mathbf{c}_i \right\|_1 + \lambda_2 \left\|\mathbf{c}_i \right\|_2^2 \right ) \; , \\
\nonumber
s.t. \enspace \mathbf{D} \geq 0 \; , \enspace \mathbf{C} \geq 0 \; .
\end{gather}

\noindent
The $\ell_1=\left\|\mathbf{c}_i \right\|_1$ penalty forces sparsity to the solution, while the $\ell_2=\left\|\mathbf{c}_i \right\|_2^2$ regularization encourages the \textit{grouping effect}~\cite{zou2005regularization}. The latter occurs when the coefficients of a regression method associated to highly correlated variables tend to be equal. In our case, the correlation is in terms of displacement directions which is induced by the local consistency of motion. This property is fundamental to let the coefficients $\mathbf{c}_i$ ($k$ row vectors of $\mathbf{C}$) expand each direction $\mathbf{d}_i$ to a spatially bounded region, ultimately letting us reproducing localized face deformations. Whereas the $\ell_1$ term grants sparsity by pushing features to zero, the quadratic term prevents sharp discontinuities to appear in the applied deformations. The positivity constraints, instead, are used to decouple opposite, concurrent movements. They can be eventually reproduced by multiplication for a negative value. This also promotes the complementarity of each learned atom~\cite{mairal2010online}.

\begin{figure}[!t]
	\centering
	\includegraphics[width=0.99\linewidth]{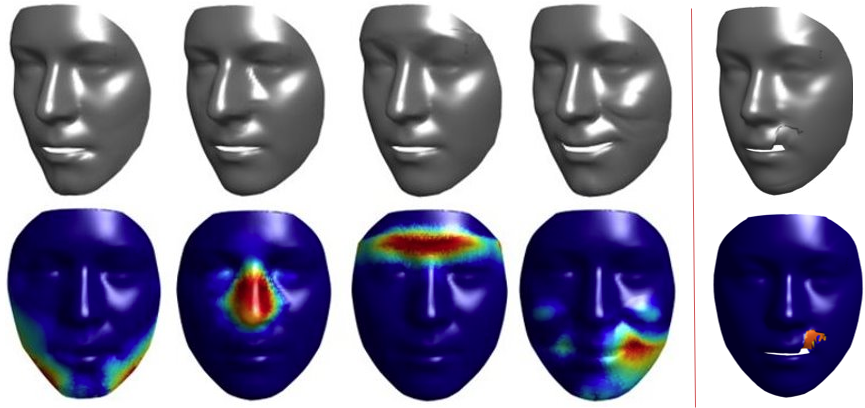}
	(a)
	\hspace{1.15cm}
	(b)
	\hspace{1.15cm}
	(c)
	\hspace{1.15cm}
	(d)
	\hspace{1.3cm}
	(e)
	\caption{Examples of learned components $\mathbf{C}$ modeling structural traits of the face related to the identity (a-b) as well as expressions (c-d). In (e) the same component as in (d) learned without $\ell_2$ regularization.}
	\label{fig:l1l2}
\end{figure}

To solve~\eqref{eq:DLSC}, we exploit the implementation in~\cite{mairal:2009} that alternates the estimation of $\mathbf{D}$ and $\mathbf{C}$, while keeping the other fixed. After convergence, we use the coefficients of the reconstruction $\mathbf{C}$ as deformation components.
Examples of the learned components are shown in Figure~\ref{fig:l1l2}. Note how identity (\eg,~nose shape, jawbone) and expression (\eg,~smiling, eyebrow raising) smooth deformations are independently reproduced and do not influence each other. The importance of the $\ell_2$ regularization to maintain a smooth surface and so obtain realistic deformations is shown in Figure~\ref{fig:l1l2}~(e). By using the $\ell_1$ penalty alone, the deformations are sparse yet discontinuous.

Finally, we observe that movable face parts present a higher degree of variability compared to other stable regions, like the forehead. This gets reflected in the magnitude of the elements of $\mathbf{D}$. In order to account for this imbalance, we compute the mean of the dictionary elements over the $N$ models for each direction $\mathbf{d}_i$, and define a weight vector $\boldsymbol{\mu} \in \mathbb{R}^k$ with elements $\mu_{j} = \frac{1}{N} \sum_{i=1}^{N} d_{i,j}, \enspace \forall j \in [1, \dots, k]$. This represents the average contribution of each direction. We will use the vector $\boldsymbol{\mu}$ to regularize the deformation of the 3DMM, balancing the contribution of each component as detailed in Section~\ref{subsec:3dmm-fitting}.
The sparse components $\mathbf{C}$, the average model $\mathbf{m}$ and the weight vector $\boldsymbol{\mu}$ constitute our \textit{Sparse and Locally-Coherent} (SLC)-3DMM.

\textbf{Discussion.} Our solution grounds on the idea that the motion correlation occurring among nearby points in the surface can be exploited to model localized facial deformations. We achieve this by explicitly analyzing the variation of each vertex coordinate independently, which is what distinguishes our solution from the previous literature. While we propose a sparse-coding formulation of the learning problem, the main novelty lies in the new way such method is applied to the data.
By comparing against recent state-of-the-art approaches under a variety of challenging conditions, we will show that a considerable improvement in the 3DMM modeling capabilities can be obtained by a motivated change in the paradigm by which training data are used. Furthermore, compared to deep-learning based approaches, the computational effort required is minimal. 

\section{Dense Semantic Annotation Transfer}\label{sec:dense_transfer}
Our goal here is that of non-rigidly deforming the SLC-3DMM to fit a target scan, while transferring its dense semantic annotation (\ie,~topology and attributes).
First, the average model $\mathbf{m}$ and a generic target shape $\mathbf{t} \in \mathbb{R}^{h \times 3}$ need to be roughly aligned. We assume the target scans are coarsely facing the camera, so that the whole face surface is visible, and the orientation is consistent to $\mathbf{m}$ (the problem of occlusions and rotations is discussed in the supplemental material). The shapes $\mathbf{t}$ are then cropped using a sphere of radius $r=95mm$ centered at the nose tip (vertex with the largest $z$ value). The shapes are zero-centered and rigidly aligned to $\mathbf{m}$ by first making the nose-tips coincident, then using rigid ICP to account for slight 3D rotation and translation shifts. This operation is performed as an initialization step. Differently from previous works, we do not use any landmarks for initialization, neither for neutral nor for expressive scans.
For the clarity of notation, we will use $\mathbf{s}$ to refer to the fitted 3DMM shape at the current iteration, \ie,~$\mathbf{s}\equiv\mathbf{m} \in \mathbb{R}^{m \times 3}$ at the beginning (see also~\eqref{eq:3dmmL}), and $\mathbf{\hat{t}}$ to refer to the coarsely aligned point-cloud.

\begin{figure}[!t]
	\centering	
	\includegraphics[width=0.99\linewidth]{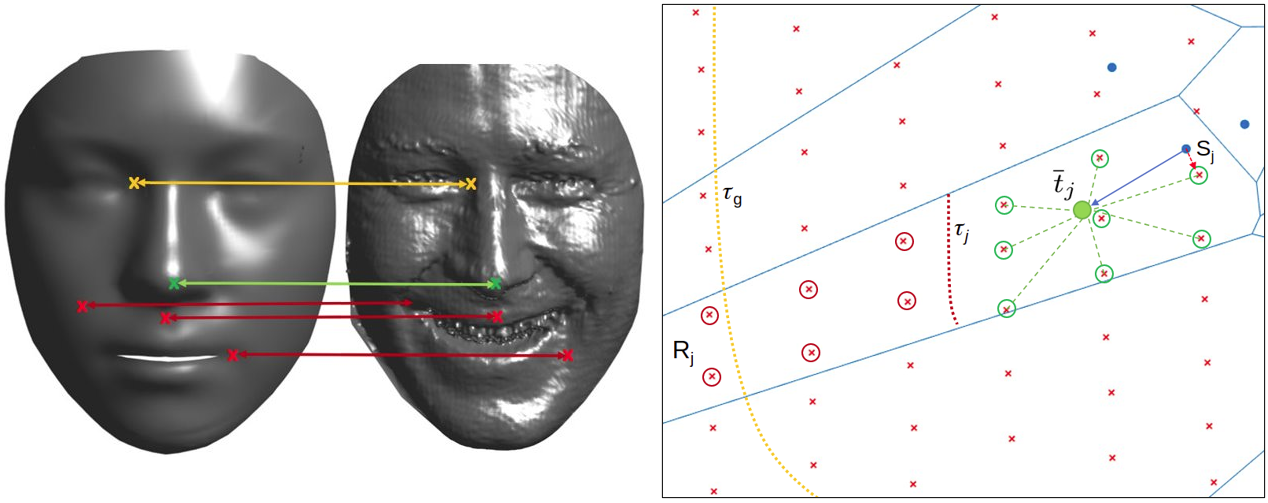}	
	\caption{\label{fig:voronoi} Large differences between template and target induce a misalignment (red=high, green=low) of local regions (left). Local Mean-Point association and outliers rejection (right): for each Voronoi region $R_j$ (blue polygon), the point $S_j$ (blue dot) is associated to the centroid $\overline{t}_j$ (green dot) of the points $\hat{t}_{R_j}$ (red crosses). The local and global rejection thresholds (red and yellow dotted lines) are iteratively updated. }
\end{figure}

\subsection{Non-rigid Fitting}\label{sec:p2p_corr}
In order to deform $\mathbf{s}$ to accurately match $\mathbf{\hat{t}}$, it is first required to establish a preliminary correspondence between them.
The naive solution to the above, like in~\cite{Gilani-dense:2018, ferrari20193dmm}, is that of first mapping each point of $\mathbf{s}$ to its nearest-neighbor (NN) in $\mathbf{\hat{t}}$, then deforming $\mathbf{s}$, and repeating these two steps until convergence.
However, when dealing with facial expressions and unseen identities, the topological difference between $\mathbf{s}$ and the target can be significant. Therefore, the initial NN correspondence can be far from optimal, likely preventing a correct shape matching. In previous works, this is solved by resorting to known pairs of landmarks that are used to initialize the deformation.

Correctly pairing each vertex of $\mathbf{s}$ to one in $\mathbf{\hat{t}}$ is the core of the whole process, as the 3DMM is deformed under the guidance of such point-to-point correspondence. In the most general case, given the possibly large difference between $\mathbf{s}$ and $\mathbf{\hat{t}}$, this mapping needs to be adjusted at small iterative steps, allowing $\mathbf{s}$ to adapt to the target surface. To this aim, we formulated a joint mean-point association and dynamic outliers rejection strategy to guide the process. In the following, we separately describe each step of our Non-Rigid Fitting (NRF) solution.

\begin{algorithm}[!t]
	\footnotesize
	\caption{{Point-to-Point Correspondence} \label{alg:p2p-assoc}}
	\KwIn{Deformed Model $\mathbf{s}$, Target Shape $\mathbf{\hat{t}}$}
	\KwOut{Re-indexed Model $\mathbf{\hat{t}}^c$}
	\medskip	
	$\mathbf{\hat{t}}^c$ = ZerosLike($\mathbf{s}$) \tcp*{Initialize $\mathbf{\hat{t}}^c$}
	$\tau_g$ = ComputeThreshold($\mathbf{s}$, $\mathbf{\hat{t}}$);\\	
	\ForEach{$s_j \in \mathbf{s}$}{
		$[R_j$, $\hat{t}_{R_j}]$ = ComputePointsVoronoiRegion($s_j$, $\mathbf{\hat{t}}$);\\
		$\tau_j$ = ComputeThreshold($s_j$, $\hat{t}_{R_j}$);\\
		$\hat{t}_{R_j}$ = RemoveOutliers($\hat{t}_{R_j}$,$\tau_g$,$\tau_j$);\\
		\uIf{$\hat{t}_{R_j} \ne \emptyset$}{
			$\overline{t}_j$ = ComputeCentroid($\hat{t}_{R_j}$);\\
			$\mathbf{\hat{t}}^c(j)$ = $\overline{t}_j$ \tcp*{Assign $\overline{t}_j$ to $\mathbf{\hat{t}}^c$ at index $j$}
			$\mathbf{s}$ = RemoveVertices($\mathbf{s}$, $j$) \tcp*{Remove $s_j$ from $\mathbf{s}$}
		}
	}	
	$[idx_{\mathbf{s}}, idx_{\mathbf{\hat{t}}}]$ = NNSearch($\mathbf{s}$,$\mathbf{\hat{t}}$)\\
	$\mathbf{\hat{t}}^c(idx_{\mathbf{s}})$ = $\mathbf{\hat{t}}(idx_{\mathbf{\hat{t}}})$;
\end{algorithm}

\subsubsection{Mean-Point Association and Outliers Rejection}\label{subsect:bidir_assoc}
We sketched the point-to-point association strategy complying with the following observations:
\begin{enumerate}
    \item Multiple vertices $\hat{t}_{R_j} = \{\hat{t}_i, \dots, \hat{t}_k\} \in \mathbf{\hat{t}}$ sharing a nearest-neighbor $s_j \in \mathbf{s}$, belong to the Voronoi region $R_j$ defined by $s_j$;
    \item Sufficiently small neighborhoods can be approximated with piecewise-planar surfaces;
    \item For large topological changes, \eg,~mouth open, the correct target point $\hat{t}_i$ to be matched is far away from the corresponding point in $\mathbf{s}$ (see Figure~\ref{fig:voronoi}~left);
    \item Scanners usually induce high-frequency surface noise that would be desirable to attenuate.
\end{enumerate}

\begin{figure*}[!t]
	\centering
	\includegraphics[width=0.9\linewidth]{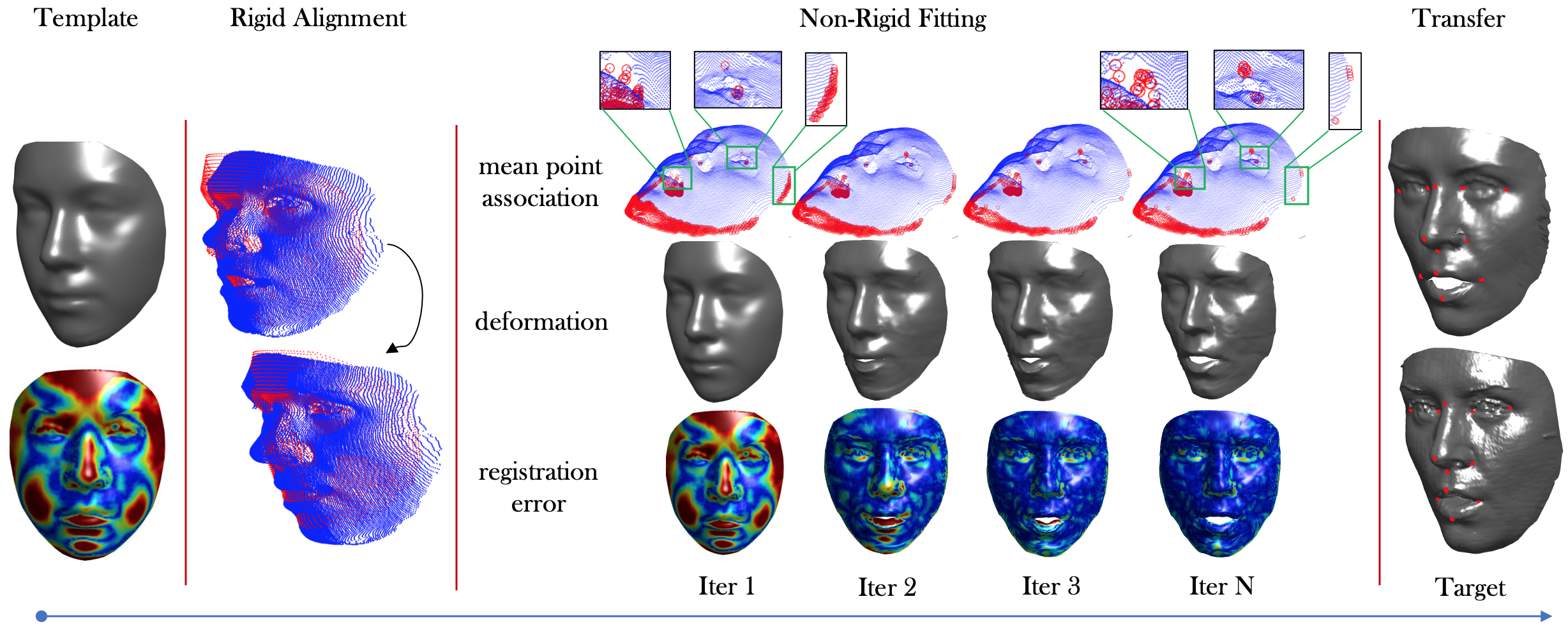}
	\caption{\label{fig:duplicates} Overview of the proposed 3DMM fitting and dense semantic transfer approach. First, the template and target are rigidly aligned. Then, $\mathbf{s}$ is iteratively deformed guided by our mean-point association procedure for point-to-point correspondence estimation. Finally, the dense annotation of the 3DMM is transferred to the target. Red points denote the detected outliers that are iteratively refined to let the template smoothly adapt to the target shape, while rejecting noisy local regions (\eg,~points inside the mouth).}
\end{figure*}

\noindent
Based on the observations above, our point-to-point association operates by mapping each vertex $s_j$ to the centroid $\overline{t}_j$ of the vertices $\hat{t}_{R_j} = \{\hat{t}_i, \dots, \hat{t}_k\}$ belonging to the Voronoi region $R_j$. The motivation for this choice is two-fold: first, it can compensate for high-frequency surface noise. More importantly, it accounts for large misalignments by avoiding getting attached to a fixed nearest-neighbor. As shown in Figure~\ref{fig:voronoi}~(left), semantically equivalent regions, \eg~mouth corner, can be highly misaligned. Using a nearest-neighbor proximity criterion can easily result inadequate, and prevent the template to adapt to large shape changes.
To handle possible outliers when computing $\overline{t}_j$, we estimate both a global and a local rejection threshold. The global threshold $\tau_g$ is computed as $\tau_g = \overline{d} + \sigma_d$, where $\overline{d}$ is the average NN-distance between $\mathbf{s}$ and $\mathbf{\hat{t}}$, and $\sigma_d$ is the standard deviation. Each local threshold $\tau_{j}$ is computed in the same way, considering points belonging to each Voronoi region $R_j$ separately. The centroids $\overline{t}_j$ are then computed after rejecting global and local outliers, and mapped at position $j$ in the re-indexed model $\mathbf{\hat{t}}^c$. The two thresholds are updated at each iteration. Finally, for each empty Voronoi region $R_j$, the association is completed using a nearest-neighbor strategy on the leftover points. At the end of the process, each point of $\mathbf{s}$ is paired. This leads to a model $\mathbf{\hat{t}}^c$ with the same semantic indexing of $\mathbf{s}$. The method is summarized in Algorithm~\ref{alg:p2p-assoc} and Figure~\ref{fig:voronoi}.

\subsubsection{Model Deformation}\label{subsec:3dmm-fitting}
The point-to-point correspondence from the previous step is exploited to refine the alignment of $\mathbf{\hat{t}}^c$ and $\mathbf{s}$. This is useful prior to deforming $\mathbf{s}$ to account for slight scale differences and encourage an iterative refinement. To this aim, we estimate a similarity transformation:
\begin{equation}
\label{eq:simTrans}
\mathbf{s} = \mathbf{\hat{t}}^c \cdot \mathbf{P} + \mathbf{1} \cdot \mathbf{T} \; ,
\end{equation}

\noindent
where $\mathbf{T} \in \mathbb{R}^{1 \times 3}$ is the 3D translation, $\mathbf{1} \in \mathbb{R}^{m \times 1}$ is the unitary column vector, and $\mathbf{P} \in \mathbb{R}^{3 \times 3}$ contains the 3D rotation and scale parameters.
$\mathbf{P}$ is found in closed-form solving the following least squares problem:
\begin{equation}
\label{eq:simTrans_sol}
\underset{\mathbf{P}}{\min}\left \| \mathbf{s} - \mathbf{\hat{t}}^c \cdot \mathbf{P} \right\|_2^2 \; .
\end{equation}

\noindent
A solution to~\eqref{eq:simTrans_sol} is given by $\mathbf{P} = \mathbf{s}^{T}\cdot (\mathbf{\hat{t}}^{cT})^\dagger$, where $\dagger$ indicates the pseudo-inverse. The translation is then recovered by aligning the barycenters, that is $\mathbf{T} = \overline{\mathbf{s}} -  \overline{\mathbf{t}^c} \cdot \mathbf{P}$.
Using $\mathbf{P}$ and $\mathbf{T}$, we re-align both $\mathbf{\hat{t}}^c$ and $\mathbf{\hat{t}}$ to $\mathbf{s}$ using~\eqref{eq:simTrans} before performing the deformation.

To deform $\mathbf{s}$, we find the optimal set of deformation coefficients $\boldsymbol{\alpha} \in \mathbb{R}^k$ so that the per-vertex distance between the two point sets is minimized. Similar to other works using a morphable model~\cite{Fan-eccv:2018, ferrari:2015, Gilani-dense:2018}, we formulate the problem as a regularized least-squares:
\begin{equation}
\label{eq:sparse-coding-b}
\underset{\boldsymbol{\alpha}}{\min}
\left \|\mathbf{\hat{t}}^c - \mathbf{s} - \mathbf{C}\boldsymbol{\alpha} \right \|_{2}^{2} + \lambda \left \| \boldsymbol{\alpha} \circ \boldsymbol{\mathbf{\mu}}^{-1} \right \|_{2} \; ,
\end{equation}

\noindent
where $\lambda$ balances the contribution of the regularization term, which enforces smoothness to let the model approximate the target at small steps. Recalling the definition of $\boldsymbol{\mu}$ in Section~\ref{subsec:build3dmm}, we regularize the deformation using the inverse $\boldsymbol{\mu}^{-1}$ so that the contribution of each component is weighed with respect to its average intensity. By pre-computing $\mathbf{X} = \mathbf{\hat{t}}^c - \mathbf{s}$, the solution is found in closed form:
\begin{equation}
\label{eq:def-coefficients-estimation}
\boldsymbol{\alpha} = \left (\mathbf{C}^T \mathbf{C} + \lambda \cdot \text{diag}(\hat{\boldsymbol{\mu}}^{-1}) \right)^{-1}\mathbf{C}^T\mathbf{X} \; ,
\end{equation}

\noindent
where $\text{diag}(\hat{\boldsymbol{\mu}}^{-1})$ denotes the diagonal matrix with vector $\hat{\boldsymbol{\mu}}^{-1}$ on its diagonal. Then, $\mathbf{s}$ is deformed applying~\eqref{eq:3dmmL}.
Finally, we estimate the per-vertex error of the deformed model as the average Euclidean distance between each vertex of $\mathbf{s}$ and its nearest-neighbor in $\mathbf{\hat{t}}$.

\begin{algorithm}[!t]	
	\footnotesize
	\caption{{Non-Rigid Fitting (NRF)} \label{alg:fitting}}
	\KwIn{Average Model $\mathbf{m}$, Sparse Components $\mathbf{C}$, Weights $\boldsymbol{\mu}$, Target Shape $\mathbf{t}$, Error Threshold $\mathbf{\tau_e}$, Iterations Limit $I_l$}
	\KwOut{Deformed Model $\mathbf{s}$}
	\medskip	
	$\mathbf{\hat{t}}$ = ICP($\mathbf{t}, \mathbf{m}$); \\
	$err$ = ComputeEuclideanError($\mathbf{\hat{t}}$,$\mathbf{m}$); \\		
	$\mathbf{s}$ = $\mathbf{m}$ , $i$ = $0$, $\delta_e$ = $\tau_e + 1$ \\
	\While{i $<$ $I_l$ $\parallel $ $\delta_e > \tau_e$}{
		$\mathbf{\hat{t}}^c$ = P2PCorrespondence($\mathbf{s}$,$\mathbf{\hat{t}}$) \tcp*{Algorithm~\ref{alg:p2p-assoc}}
		$[\mathbf{\hat{t}}^c, \mathbf{\hat{t}}]$ = SimilarityTransform($\mathbf{\hat{t}}^c$,$\mathbf{\hat{t}}$,$\mathbf{s}$) \tcp*{Eq.~\eqref{eq:simTrans},~\eqref{eq:simTrans_sol}}
		$\mathbf{s}$ = SLCFitting($\mathbf{\hat{t}^c},\mathbf{s}, \mathbf{C}, \boldsymbol{\mu}$)
		\tcp*{Eq.~\eqref{eq:sparse-coding-b},~\eqref{eq:def-coefficients-estimation}}
		$e$ = ComputeEuclideanError($\mathbf{\hat{t}}$,$\mathbf{s}$); \\
		$\delta_e$ = $err - e$;\\
		$err$ = $e$;\\
		$i = i + 1$;}
\end{algorithm}

The NRF procedure is repeated until the error between subsequent iterations is above some threshold $\tau_{e}$, or a maximum number of iterations is reached (see Algorithm~\ref{alg:fitting}).

\textbf{Discussion:} A good strategy to establish a preliminary correspondence between points in $\mathbf{\hat{t}}$ and $\mathbf{s}$ and correctly guide the deformation is of utmost importance. While the template deformation is achieved using quite a standard solution, the main novelty of our NRF procedure lies in the way such correspondences are found. The overall idea is exemplified in Figure~\ref{fig:voronoi}~(right). Each point $s_j$ is provided with the centroids $\overline{t}_j$ as target for the deformation. At the next iteration, $\mathbf{s}$ is deformed, and the new location of each $s_j$ will define new Voronoi regions $R_j$, and consequently new rejection thresholds and centroids. This is the core of our proposal, which lets $\mathbf{s}$ iteratively adapt to the global shape of $\mathbf{\hat{t}}$ at small steps, until the two shapes have, on average, minimum distance. Differently, using a nearest-neighbor criterion as adopted by previous works \eg~\cite{Amberg-nicp:2007, Gilani-dense:2018}, cannot effectively compensate for large misalignments.

\subsection{Transfer the Semantic Annotation}\label{subsec:annotation_transfer}
Once the 3DMM is fit to the target shape, we can transfer the semantic labeling of $\mathbf{s}$ to $\mathbf{\hat{t}}$. To this end, each point of $\mathbf{s}$ must be associated to a distinct point in $\mathbf{\hat{t}}$, so that the latter has the same semantic indexing of the 3DMM.
This is achieved by computing, for each vertex in $\mathbf{s}$, its $k$-nearest neighbors in $\mathbf{\hat{t}}$. If multiple points in $\mathbf{s}$ 
are mapped to the same point $\mathbf{\hat{t}}_k$, the closest one is assigned to $\mathbf{\hat{t}}_k$. The remaining points are then mapped to the ($k$-1)-nearest neighbors following the same scheme. This strategy leads to smoother meshes, where no vertices of $\mathbf{s}$ are collapsed into the same point. The process ends when each point in $\mathbf{s}$ has been paired to a unique point in $\mathbf{\hat{t}}$, leading to a re-indexed model $\mathbf{\hat{t}}'$ with the same semantic labeling of $\mathbf{s}$. Figure~\ref{fig:duplicates} summarizes the overall fitting and semantic transfer approach.

\section{Experimental Results}\label{sec:experimental-results}
Assessing a dense correspondence algorithm is a complex task since no solid definition of ground-truth does exist. We carried out an extensive experimental validation to showcase the quality of our solution under different perspectives.
First, we compare SLC against other state-of-the-art face modeling approaches (Section~\ref{subsec:ablation}). Following the standard practice, we then evaluate our NRF approach in the task of 3D landmark detection, also in comparison with state-of-the-art solutions (Section~\ref{subsec:nrf}). Finally, we validate the accuracy of our registration approach by densely aligning scans from heterogeneous datasets and constructing a large-scale 3DMM from them. Results for different datasets are given in Section~\ref{subsec:3dmm-6K} and Section~\ref{subsec:hls_vs_bufe} also in comparison with other registration methods.

\subsection{Datasets}\label{subsec:datasets}
Overall, we have used five different 3D face datasets.
The BU-3DFE~\cite{yin20063d} and the CoMA~\cite{ranjan2018generating} datasets were used to explore the performance of the proposed SLC learning solution.
To evaluate our dense registration algorithm, instead, we used the Face Recognition Grand Challenge (FRGCv2.0)~\cite{phillips:2005}, the Bosphorus~\cite{savran:2008}, and the FaceWarehouse~\cite{cao2013facewarehouse} datasets that cover a wide range of different characteristics for 3D faces.

\textbf{BU-3DFE} -- It comprises $100$ subjects in neutral plus six expressions, ranging from low to exaggerated intensity level. For this work, we employed the fully registered version of the data as obtained in~\cite{ferrari:2015}. This subset includes $1,779$ scans, each with $6,704$ vertices.

\textbf{CoMA} -- It consists of $12$ subjects, each one performing $12$ extreme and asymmetric expressions. Each expression comes as a sequence of fully-registered meshes with $5,023$ vertices. Each sequence is composed of $140$ meshes on average, for a total of $20,466$ scans.

\textbf{FRGCv2.0} -- It includes $4,007$ scans of $466$ individuals collected in two separate sessions. Approximately, 40\% of the scans show slight spontaneous expressions. On average, the FRGCv2.0 scans have $35K$ vertices in the face region.

\textbf{Bosphorus} -- This dataset comprises $4,666$ high-resolution scans of $105$ individuals. There are up to $54$ scans per subject, which include prototypical expressions and facial action unit activations. The raw scans of Bosphorus have an average of $30K$ vertices on the face region. This dataset also contains rotated and occluded scans that we did not use. In fact, in both the cases, the problem of dense correspondence is ambiguous. An evaluation on these subsets is reported in the supplemental material.

\textbf{FaceWarehouse} -- This dataset comprises $3,000$ Kinect RGB-D sequences of $150$ individuals, aged 7-80. For each individual, $20$ sequences are captured that include the neutral expression plus $19$ natural expressions such as mouth-opening, smiling, kissing,~\etc~Only the first RGB-D frame of each sequence has been released for public use. The point-clouds extracted from the frames have about $15K$ vertices.

\begin{figure*}[!t]
	\centering
	\includegraphics[width=0.99\linewidth]{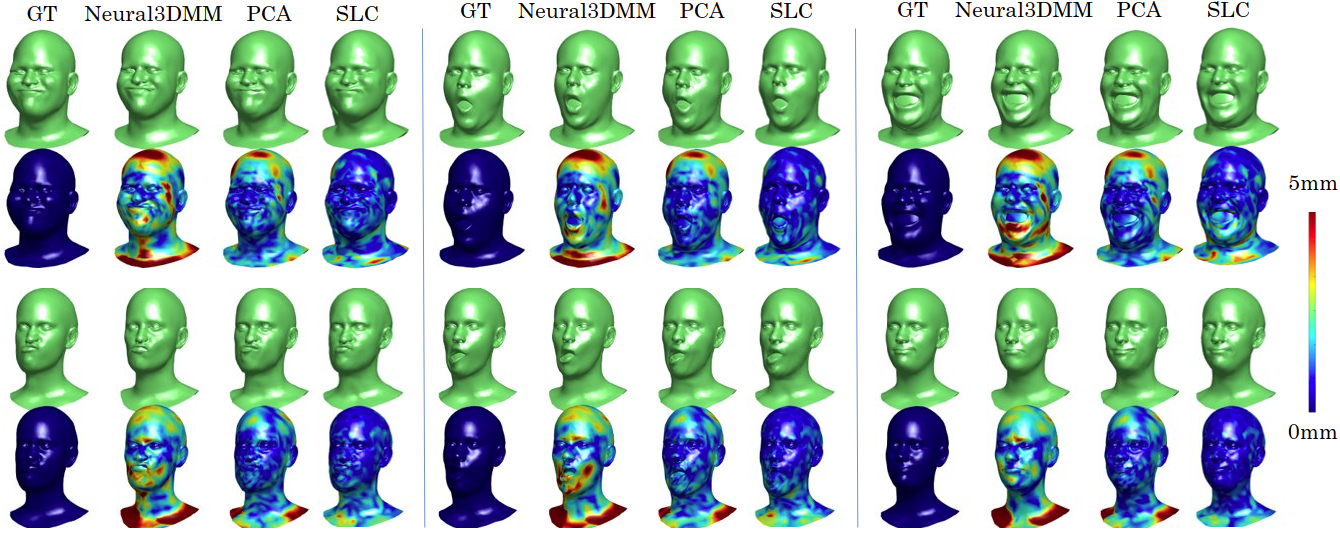}	
	\caption{\label{fig:fitExCOMA} CoMA dataset: Qualitative comparison of some reconstructions obtained with the Neural3DMM~\cite{bouritsas2019neural}, PCA and our SLC solution (better viewed in digital). Error heatmaps show that SLC results in more accurate reconstructions.}
\end{figure*}

\subsection{SLC Components Learning}\label{subsec:ablation}
The major trait that differentiates the SLC solution with respect to previous 3DMM literature is the way in which we interpret the training data. With our setup, independently from the number $N$ of face scans, $3m$ samples are used for training, being $m$ the number of vertices. We can thus build powerful overcomplete representations with $k > N$ components, capturing a larger variety of patterns in the data~\cite{lee2007efficient}. We will show this leads to remarkable generalization capabilities with respect to state-of-the-art solutions, mostly for smaller training sets with limited variability.

We report results obtained on the CoMA and BU-3DFE datasets. We compare against the DL-3DMM~\cite{ferrari:2015}, and the standard PCA-3DMM, using the parameters suggested in~\cite{ferrari:2015}. Furthermore, we also compare against two recent state-of-the-art methods for learning 3D face representations, namely, the Convolutional Mesh Autoencoder (CoMA)~\cite{ranjan2018generating}, and the Neural3DMM~\cite{bouritsas2019neural} methods. We trained both using the publicly available code with the suggested settings. Given the fixed topology of the meshes, when using PCA, DL and SLC, for each shape in the test set we estimate the deformation coefficients $\mathbf{\alpha}$ using~\eqref{eq:def-coefficients-estimation}, and deform the average model using~\eqref{eq:3dmmL}. When using CoMA and Neural3DMM, test shapes are provided as input to the autoencoders, and the error is computed on the reconstructed shapes. 
For all the subsequent experiments, we set $\lambda_1 = 1$ and $\lambda_2 =1$ for learning the SLC components. Errors are reported in terms of average per-vertex Euclidean distance.

\begin{figure}[!t]
	\centering
	\includegraphics[width=0.99\linewidth]{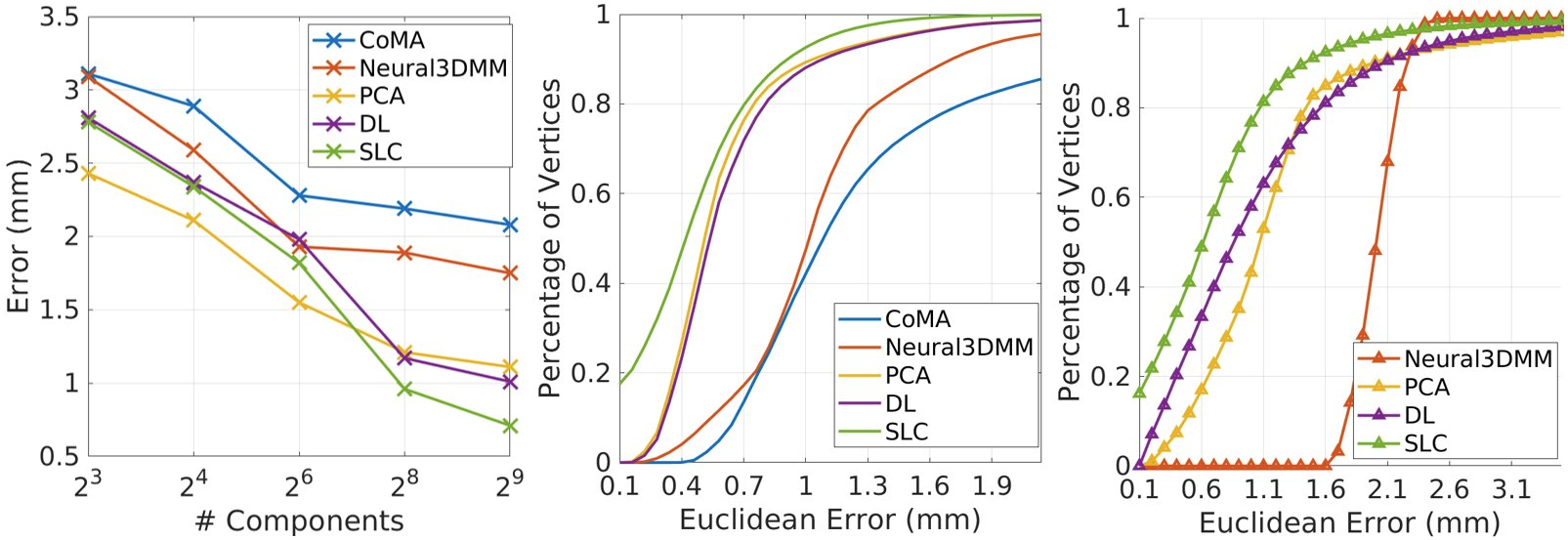}
	(a)
	\hspace{2.3cm}
	(b)	
	\hspace{2.3cm}
	(c)
	\caption{\label{fig:ablationCOMA} CoMA dataset: Ablation study against CoMA~\cite{ranjan2018generating}, Neural3DMM~\cite{bouritsas2019neural}, DL~\cite{ferrari:2015} and PCA. The effect of varying the latent vector size, \ie,~number of components (a), cumulative per-vertex error distribution using $z=k=512$ (b), and learning the models on a reduced training set ($360$ scans instead of $\approx19,000$) (c) are reported.}
\end{figure}

\subsubsection{CoMA Database}\label{subsubsec:slc-CoMA}
The CoMA database contains 12 identities, each performing 12 sequences of complex, non-standard expressions. To fully evaluate the generalization ability of the models to new identities that are not included in the training data, we perform experiments using the ``\textit{Identity}'' data splitting as defined in~\cite{ranjan2018generating}. With this protocol,~11 identities are used for training, while testing is performed on the leftover subject. Results are averaged over a 12-fold cross validation.

Firstly, we evaluate the effect of changing the size of the latent vector $z$ in~\cite{bouritsas2019neural, ranjan2018generating}, in comparison with the number of deformation components $k$ used for PCA, DL and SLC. Figure~\ref{fig:ablationCOMA}~(a) reports the mean per-vertex Euclidean distance as a function of the number of components, either $z$ or $k$.
From Figure~\ref{fig:ablationCOMA}~(a), it turns out that both~\cite{ranjan2018generating} and~\cite{bouritsas2019neural} do not retain satisfactory precision on samples whose identity is not in the training, disclosing a significant weakness of such methods. Given the sparse nature of our deformation components, when using a small $k$ (up to 64 components) PCA performs best; however, the reconstruction error is rather large, about 1.5mm. When increasing $k$, our SLC outperforms all the compared solutions.
Figure~\ref{fig:ablationCOMA}~(b) reports the cumulative, per-vertex error distribution obtained with $k=512$. The significantly improved fitting precision obtained with SLC comes out clearly from the results.
Finally, Figure~\ref{fig:ablationCOMA}~(c) showcases the robustness of SLC in the case of limited training data. We performed the same experiment as in Figure~\ref{fig:ablationCOMA}~(b), but using only 360 scans, \ie,~2\% of the training data. The accuracy of all the compared solutions drops significantly, while SLC attains a similar result to that obtained on the full training set. The percentage of vertices with an error lower than 1mm drops from $92\%$ to $79\%$, while for Neural3DMM, PCA and DL, it drops down to $0\%$, $42\%$, and $59\%$, respectively (the CoMA~\cite{ranjan2018generating} method performed significantly worse and results were not reported). Some examples are shown in Figure~\ref{fig:fitExCOMA}.

\begin{figure}[!t]
	\centering
	\includegraphics[width=0.95\linewidth]{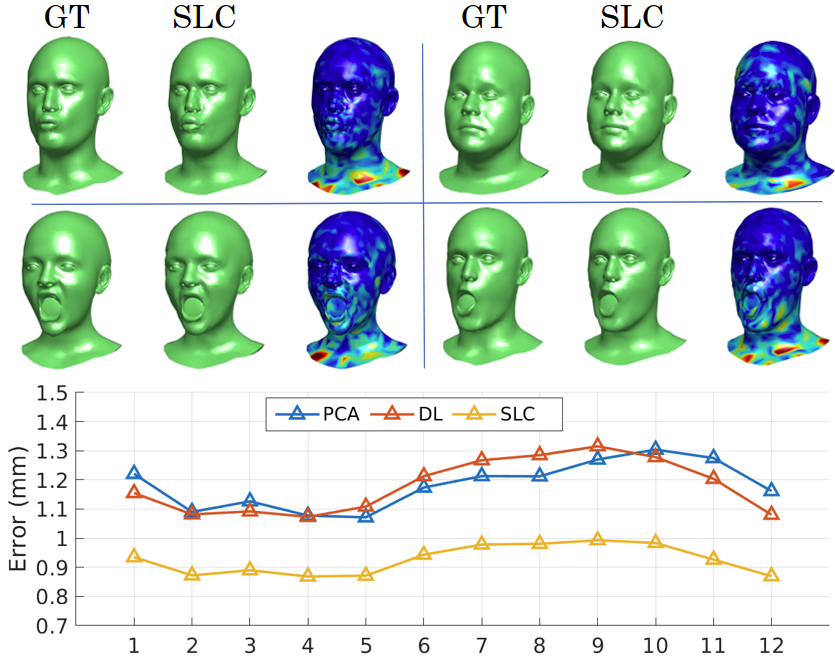}	
	\caption{\label{fig:extrapCOMA} CoMA dataset: Testing on unseen identities and expressions. Top row: SLC reconstructed samples. Bottom row: average per-vertex Euclidean error is reported for each of the 12 expressions ($x$-axis).}
\end{figure}

In Figure~\ref{fig:extrapCOMA}, we compare the above methods under a more challenging setting, where each one of the 12 expressions is in turn excluded from the training set. So, we train on 11 identities and 11 expressions, and test on the leftover expression sequence of the excluded identity. In total, we perform a 12-fold cross validation on the expressions, averaging over 5 different subjects. Remarkably, SLC outperforms the compared solutions, reconstructing the unseen samples, both identity and expression, with high precision (error $< 1$mm). We did not report results for CoMA and Neural3DMM as they were significantly worse. The qualitative examples in Figure~\ref{fig:extrapCOMA}~(top row) show that the reconstructed shapes are faithful to the ground-truth both in terms of expression and head shape.

\begin{figure}[!t]
	\centering
	\includegraphics[width=0.99\linewidth]{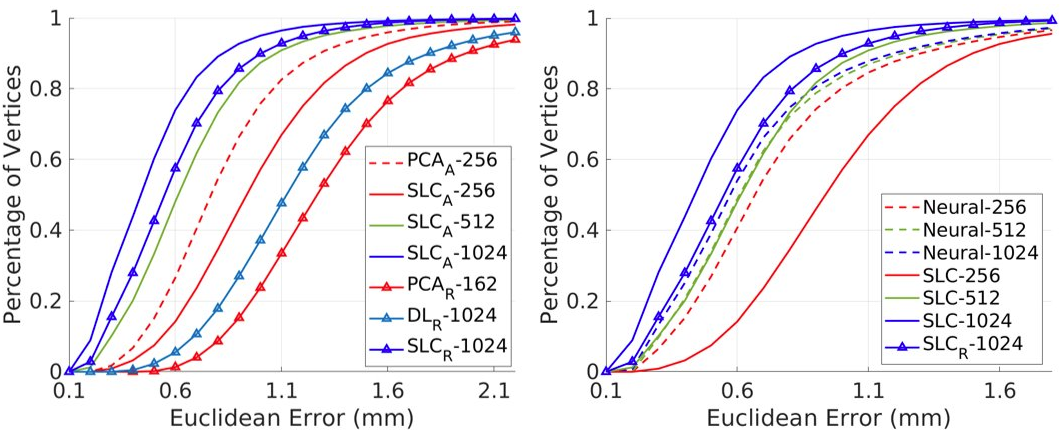}
	(a)
	\hspace{3.9cm}
	(b)
	\caption{\label{fig:ablationTrain} BU-3DFE dataset: Cumulative per-vertex error distribution comparing against (a) PCA and DL, (b) Neural3DMM. Legends report ``Method$_T$ - $k$'' where the subscript $T$ refers either to the complete training set ``A'', or the reduced training set ``R'' (10\% of the samples). $k$ is the number of components.}
\end{figure}

\subsubsection{BU-3DFE}\label{subsubsec:slc-trainSize}
For this experiment, we split the $1,779$ fully registered scans of the BU-3DFE from~\cite{ferrari:2015}, into train and test sets. The test set is composed of 10 out of 100 identities (5 males and 5 females), for a total of 162 scans, comprising neutral as well as slight and strong expressions.
In Figure~\ref{fig:ablationTrain}~(a), we compare SLC against PCA and DL. Our aim here is to show that using the SLC solution, significantly more accurate reconstructions can be obtained even in the case a very little number of training scans is available. The dashed red curve is obtained with 256 PCA components (roughly 99.5\% of the variance) and serves us as a baseline. 
Given the larger data variability carried by the training identities, here the performance of PCA is better than SLC (solid lines) when using $k=256$ components. However, by increasing the number of SLC components, a significantly more accurate level of reconstruction can be achieved. Again, the actual value of our solution comes out in the case of reduced training sets. The three dotted curves (PCA$_R$, DL$_R$ and SLC$_R$) refer to using a reduced training set of 162 scans, obtained by considering the 10\% of the training samples. In this case, the number of PCA components cannot exceed 162, \ie,~the number of samples, and the performance is dramatically reduced. With the DL-3DMM, we can use a larger number of dictionary atoms; however, they represent a redundant linear combination and do not bring any significant improvement. Thus, the DL-3DMM performs better than PCA, but still the results are unsatisfactory. With our proposed SLC, instead, we can build powerful overcomplete representations, which result in a very high reconstruction accuracy. Similarly to the outcomes reported in Figure~\ref{fig:ablationCOMA}~(c), the drop of accuracy for SLC with respect to the full training set is relatively small.

A very similar behavior is observed for Neural3DMM in Figure~\ref{fig:ablationTrain}~(b). The accuracy of the Neural3DMM when using a latent vector of either 256, 512 or 1024 elements is very similar, indicating a saturation point of the performance. With SLC, the accuracy instead keeps increasing substantially. We also report the accuracy obtained with SLC on the reduced training set (SLC$_R$). Remarkably, results are still slightly better than those of Neural3DMM, despite SLC being learned on 162 samples only. In this scenario, the merit of our learning solution is clear.

\subsubsection{Discussion}
Experiments on the CoMA and BU-3DFE datasets, highlighted the value and the advantages of the proposed SLC-3DMM. We showed that its generalization capabilities are way more pronounced with respect to both standard and state-of-the-art solutions, mostly when the training data are scarce. With just a handful of training samples, we can still surpass the accuracy of previous models by increasing the number of components $k$. Differently from the compared methods, the reason why the SLC performance tends not to saturate for larger $k$ can be found recalling~\eqref{eq:DLSC}. When increasing $k$ and fixing both $\lambda_1$ and $\lambda_2$, the regularization terms imposed to the components $\mathbf{c}_i$ force a larger sparsity. This implies smaller regions are deformed, ultimately allowing us to model finer, localized yet complementary surface areas. To provide a coarse measure, with $k=64$, the $84.4\%$ of the entries of $\mathbf{c}$ are zero; with $k=512$, the percentage increases to $96.3\%$. Compared with DL-3DMM, which used a similar formulation to learn the components, we achieve a remarkable improvement. We believe this clearly indicates the value of our intuition that can open the way to new and alternative learning strategies.

On the other hand, a drawback is that such sparsity makes it necessary to retain a sufficient number of components for accurately modeling the entire surface. In fact, each component acts on a spatially bounded region, the extent of which cannot increase excessively. This because of the local consistency of motion property. When the training data are sufficiently variegated and numerous, previous methods such as PCA obtain better reconstructions with smaller $k$. However, we observe that SLC outperforms all the compared methods when $k$ increases. Moreover, collecting such variegated data is not straightforward; as demonstrated by the experiments on CoMA, a large number of samples does not imply large variability. So, extending the modeling capabilities to very different yet unseen data becomes crucial, which was the main goal that motivated the development of the SLC method.
In the following, we will show that when the target scans of unknown topology to be reconstructed deviate largely from the training data, the SLC components bring a valuable improvement.

\begin{figure*}[!t]
	\centering
	\includegraphics[width=0.99\linewidth]{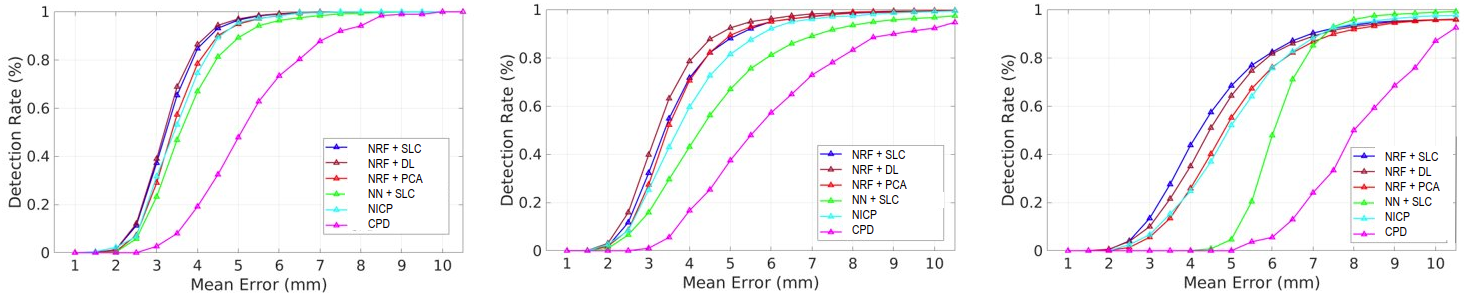}
	(a) FRGCv2.0
	\hspace{3.5cm}
	(b) Bosphorus
	\hspace{3.5cm}
	(c) FaceWarehouse
	\caption{Cumulative landmark localization error distributions for (a) FRGCv2.0, (b) Bosphorus, (c) FaceWarehouse.}
	\label{fig:cumlmErr}
\end{figure*}

\subsection{NRF and Dense Registration}\label{subsec:nrf}
In this section, we evaluate our dense registration method against previous techniques in a landmark detection task. Since our proposal consists of a combination of a novel method for learning the 3DMM components (SLC) and a model fitting algorithm (NRF), we will compare against methods in both the categories.

\textbf{Parameter Settings} -- For the following experiments, the error threshold $\tau$ and the maximum number of iterations in Algorithm~\ref{alg:fitting} were set to $0.01$ and $30$, respectively. The regularization parameter $\lambda$ in~\eqref{eq:def-coefficients-estimation} was set to 1. The SLC parameters were set to $k=50$, $\lambda_1=1$ and $\lambda_2=1$. More details can be found in the supplementary material.

\textbf{Computational Time} -- Table~\ref{tab:comp-times}~(left) reports the computational time of each step of the dense registration algorithm, evaluated on a machine with a Xeon CPU (2.6GHZ, single thread). The average time of the complete pipeline depends on the number of NRF iterations and can vary from $\approx$ 10s (1 iteration) to $\approx$ 140s (worst case scenario of 30 iterations). On average, 8-10 iterations are sufficient for convergence, taking approximately $40-50$ seconds. Note that all the operations are highly parallelizable, and code optimizations can further reduce the computational times. In Table~\ref{tab:comp-times}~(right) the average times for learning the deformation components are reported. Note that PCA, DL~\cite{ferrari:2015} and the proposed SLC run in CPU, while for training CoMA~\cite{ranjan2018generating} and the Neural3DMM~\cite{bouritsas2019neural} we used a Titan-Xp GPU. The reported times refer to learning models with $k=z=512$ on the training splits of the CoMA database ($\approx 18,000$ scans).

\begin{table}[!t]
	\caption{\label{tab:comp-times} Average times for dense registration and models learning.}
	\begin{center}
		
		\begin{tabular}{llc}
			\multicolumn{3}{c}{\textbf{Dense Registration}} \\
			\hline
			\multicolumn{2}{l}{ \rule{0pt}{2ex} \textbf{Step}}  & \textbf{Time (sec)} \\
			\hline
			Init. & ICP & 0.9 \\
			\hline
			\multirow{2}{*}{NRF} & Sect.~\ref{subsect:bidir_assoc} & 4.5 \\
			& Sect.~\ref{subsec:3dmm-fitting} & 0.03 \\				
			\hline
			Transfer & Sect.~\ref{subsec:annotation_transfer} & 5.0 \\				
			\hline			 
			\multicolumn{2}{l}{ \rule{0pt}{3ex} \textbf{Total}} & \textbf{Time (sec)}\\
			\hline
			\multicolumn{2}{l}{Proposed} & 10-140 \\
			\multicolumn{2}{l}{Fan~\etal~\cite{Fan-cvpr:2019}} & 100 \\
			\multicolumn{2}{l}{Fan~\etal~\cite{Fan-eccv:2018}} & 160 \\
			\multicolumn{2}{l}{NICP~\cite{Amberg-nicp:2007}} & 80 \\
			\hline
		\end{tabular}
		\quad
		\raisebox{7.5mm}{
		\begin{tabular}{lc}
			\multicolumn{2}{c}{\textbf{Model Learning}} \\
			\hline
			\rule{0pt}{2ex} \textbf{Method} & \textbf{Time}\\
			\hline	
			PCA & 4.49 sec\\
			DL~\cite{ferrari:2015} & 14.65 sec \\
			SLC & 6.10 sec \\		
			\hline
			CoMA~\cite{ranjan2018generating} & 9 hours\\
			Neural~\cite{bouritsas2019neural} & 8 hours \\
			\hline		
		\end{tabular}
	}
	\end{center}
\end{table}

\subsubsection{Landmarks Detection}\label{subsubsec:landmark-detection}
First, to assess the contribution of the mean-point association and outliers rejection in the NRF procedure, we substituted Algorithm~\ref{alg:p2p-assoc} with the standard nearest-neighbor strategy for point-to-point correspondence. This solution is called NN+SLC in the following. Further, we compared against two standard non-rigid point-cloud matching techniques: Coherent Point Drift (CPD)~\cite{myronenko2010point} and Non-Rigid ICP (NICP)~\cite{Amberg-nicp:2007}.
Finally, we explored the behavior of the classic PCA-based 3DMM and the DL-3DMM~\cite{ferrari:2015}, when used in place of the proposed SLC in conjunction with NRF. In the following, these two solutions are called, respectively, NRF+PCA and NRF+DL.
For the PCA and DL variants, we chose the best configurations as reported in~\cite{ferrari:2015}.
To evaluate the error, we first manually labeled a landmark configuration on the average model $\mathbf{m}$, obtaining a set of indices $L_{idx}$. The error is computed between the ground-truth annotations in $\mathbf{\hat{t}}$ as provided by~\cite{Creusot-ijcv:2013, cao2013facewarehouse}, and our transferred landmarks $\mathbf{\hat{t}'}(L_{idx})$.

\begin{table*}[!t]
	\caption{\label{tab:sota-landmarks-frgc} FRGCv2.0: Landmark localization error (mm). \textbf{Ex/En}-outer/inner eye corner, \textbf{N}-nose bridge, \textbf{Prn}-nose-tip, \textbf{Sn}-nasal base, \textbf{Ac}-nose corner, \textbf{Ch}-mouth corner, \textbf{Ls/Li}-upper/lower lip. Landmarks for Ex-En-Ch-Ac have been averaged. Best results in bold, second best in italic.}
	\begin{center}

		\scalebox{0.93}{
			\begin{tabular}{lcccccccccc}
				\hline
				\textbf{Method} & \textbf{Ex} & \textbf{En} & \textbf{N} & \textbf{Prn} & \textbf{Sn} & \textbf{Ac} & \textbf{Ch} & \textbf{Ls} & \textbf{Li} & \textbf{Avg} \\
				\hline
				Creusot \textit{et al.}~\cite{Creusot-ijcv:2013} & $5.9 \pm 3.1$ & $4.3 \pm 2.2$ & $4.2 \pm 2.1$ & $3.4 \pm 2.0$ & $3.7 \pm 3.1$ & $4.8 \pm 3.6$ & $5.6 \pm 3.5$ & $4.2 \pm 3.2$ & $5.5 \pm 3.3$ & $4.6 \pm 2.9$\\
				BFM~\cite{paysan20093d} & $\mathbf{2.5 \pm 2.2}$ & $\mathit{2.7 \pm 2.2}$ & $3.2 \pm 2.2$ & $2.3 \pm 2.0$ & $3.8 \pm 3.6$ & $8.3 \pm 2.9$ & $\mathbf{2.6 \pm 2.9}$& $\mathbf{2.6 \pm 2.2}$ & $3.8 \pm 3.7$ & $3.5 \pm 2.7$ \\
				Sukno \textit{et al.}~\cite{sukno20143} & $4.7 \pm 2.7$ & $3.6 \pm 1.7$ & $2.5 \pm 1.6$ & $2.3 \pm 1.7$ & $2.7 \pm 1.1$ & $\mathbf{2.6 \pm 1.4}$ &$\mathit{3.9 \pm 2.8}$ & $3.3 \pm 1.8$ & $4.6 \pm 3.4$ & $3.3 \pm 2.0$\\
				K3DM$_{BU}$~\cite{Gilani-dense:2018} &
				$\mathbf{2.5 \pm 2.2}$ &
				$\mathbf{2.4 \pm 1.9}$ &
				$2.8 \pm 1.8$ &
				$2.6 \pm 1.6$ &
				$3.6 \pm 1.9$ &
				$6.1 \pm 2.7$ &
				$4.2 \pm 3.1$ &
				$2.9 \pm 3.4$ &
				$4.6 \pm 2.9$ &
				$3.5 \pm 2.2$\\
				NRF+PCA &
				$ 4.0 \pm 1.8 $ &
				$ 3.3 \pm 1.4 $ &
				$ \mathit{2.1 \pm 1.6 }$ &
				$ \mathit{1.9 \pm 1.0} $ &
				$ \mathbf{1.4 \pm 1.4} $ &
				$ 3.5 \pm 1.2 $ &
				$ 4.5 \pm 2.2 $ &
				$ 2.8 \pm 1.4 $ &
				$ 3.6 \pm 2.1 $ &
				$ 3.0 \pm 1.6 $  \\
				NRF+DL~\cite{ferrari:2015} &
				$ 3.9 \pm 1.8 $ &
				$ 3.1 \pm 1.3 $ &
				$ \mathit{2.1 \pm 1.6} $ &
				$ \mathbf{1.8 \pm 1.0} $ &
				$ \mathbf{1.4 \pm 1.3} $ &
				$ 3.1 \pm 1.1 $ &
				$ 4.1 \pm 2.0 $ &
				$ \mathit{2.7 \pm 1.4} $ &
				$ \mathbf{3.5 \pm 2.0} $ &
				$ \mathit{2.9 \pm 1.6} $  \\
				NRF+SLC &
				$ \mathit{3.8 \pm 1.8} $ &
				$ 3.0 \pm 1.3 $ &
				$ \mathbf{1.8 \pm 1.5} $ &
				$ \mathit{1.9 \pm 1.0} $ &
				$ \mathit{1.5 \pm 1.5} $ &
				$ \mathit{2.9 \pm 1.2} $ &
				$ 4.2 \pm 2.2 $ &
				$ 2.8 \pm 1.5 $ &
				$ \mathit{3.6 \pm 2.1} $ &
				$ \mathbf{2.8 \pm 2.1} $  \\
				\hline
			\end{tabular}
		}
	\end{center}
\end{table*}

\begin{figure}[!t]
	\centering
	\includegraphics[width=0.9\linewidth]{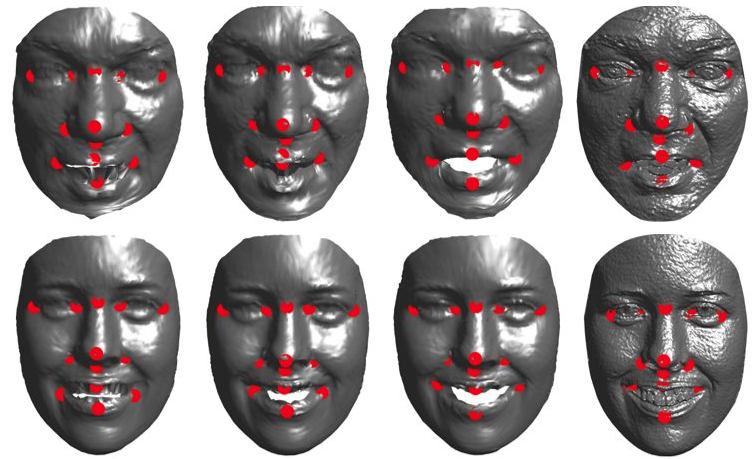}
	\hspace{1.4cm}
	(a)
	\hspace{1.4cm}
	(b)
	\hspace{1.4cm}
	(c)
	\hspace{1.4cm}
	(d)
	\caption{\label{fig:com_nn_mp} Re-indexed models $\mathbf{t}'$ with transferred landmarks $\mathbf{t}'(L_{idx})$ obtained using NICP~\cite{Amberg-nicp:2007} (a), the standard Nearest-Neighbor (b), and our proposed mean point association (c). The ground-truth scans are shown in (d). Our solution maintains a semantic consistency in case of large expressions and noise, especially in the mouth region.}
\end{figure}

\begin{table*}[t]
	\caption{\label{tab:sota-landmarks-bosph} Bosphorus: Landmarks localization error (mm). \textbf{Ex/En}-outer/inner eye corner, \textbf{N}-nose bridge, \textbf{Prn}-nose-tip, \textbf{Sn}-nasal base, \textbf{Ac}-nose corner, \textbf{Ch}-mouth corner, \textbf{Ls/Li}-upper/lower lip. Landmarks for Ex-En-Ch-Ac have been averaged. Best results in bold, second best in italic.}
	\begin{center}
		\scalebox{0.9}{
			\begin{tabular}{lccccccccccc}
				\hline
				\textbf{Method}  & \textbf{Ex} & \textbf{En} & \textbf{N} & \textbf{Prn} & \textbf{Sn} & \textbf{Ac} & \textbf{Ch} & \textbf{Ls} & \textbf{Li} & \textbf{Avg} \\
				\hline
				Cruesot \textit{et al.}~\cite{Creusot-ijcv:2013}  &
				$ 5.2 \pm \star $ &
				$ 4.6 \pm \star$ &
				$ 6.3 \pm \star$ &
				$ 4.5 \pm \star$ &
				$ 15.2 \pm \star$ &
				$ 4.1 \pm \star$ &
				$ 6.0 \pm \star$ &
				$ 6.5 \pm \star$ &
				$ 6.5 \pm \star$ &
				$ 6.3 \pm \star$ \\
				BFM~\cite{paysan20093d} &
				$ \mathbf{3.6 \pm \star} $ &
				$ \mathit{2.7 \pm \star} $ &
				$ 2.2 \pm \star$ &
				$ 2.9 \pm \star$ &
				$ 3.6 \pm \star$ &
				$ 4.0 \pm \star$ &
				$ 5.9 \pm \star$ &
				$ 4.0 \pm \star$ &
				$ 6.5 \pm \star$ &
				$ 3.9 \pm \star$ \\
				Sukno \textit{et al.}~\cite{sukno20143}  &
				$ 5.1 \pm \star$ &
				$ 2.8 \pm \star$ &
				$ 2.2 \pm \star$ &
				$ 2.3 \pm \star$ &
				$ 2.8 \pm \star$ &
				$ \mathit{3.0 \pm \star} $ &
				$ 6.1 \pm \star$ &
				$ 5.3 \pm \star$ &
				$ 5.3 \pm \star$ &
				$ 4.3 \pm \star$ \\
				K3DM$_{BO}$~\cite{Gilani-dense:2018} &
				$ \mathbf{3.6} \pm \star$ &
				$ \mathbf{2.5} \pm \star$ &
				$ 2.3 \pm \star$ &
				$ 2.8 \pm \star$ &
				$ 2.3 \pm \star$ &
				$ \mathbf{2.7} \pm \star$ &
				$ \mathit{4.9 \pm \star} $ &
				$ 3.3 \pm \star$ &
				$ 5.0 \pm \star$ &
				$ 3.3 \pm \star$ \\
				NRF+PCA  &
				$ 4.1 \pm 2.2 $ &
				$ 3.0 \pm 1.5 $ &
				$ 2.0 \pm 1.2 $ &
				$ \mathbf{2.1 \pm 1.4} $ &
				$ \mathit{1.8 \pm 1.9} $ &
				$ 3.8 \pm 1.6 $ &
				$ 5.1 \pm 2.8 $ &
				$ 3.3 \pm 1.8 $ &
				$ \mathbf{4.0 \pm 3.0} $ &
				$ 3.2 \pm 1.9$ \\
				NRF+DL~\cite{ferrari:2015}  &
				$ \mathit{3.8 \pm 2.0} $ &
				$ \mathit{2.7 \pm 1.4} $ &
				$ \mathit{1.7 \pm 1.1} $ &
				$ \mathbf{2.1 \pm 1.3} $ &
				$ \mathbf{1.8 \pm 1.7} $ &
				$ 3.4 \pm 1.4 $ &
				$ \mathbf{4.7 \pm 2.6} $ &
				$ \mathbf{2.9 \pm 1.7} $ &
				$ \mathbf{4.0 \pm 3.1} $ &
				$ \mathbf{3.0 \pm 1.9} $ \\
				NRF+SLC  &
				$ 3.9 \pm 2.2 $ &
				$ 2.9 \pm 1.5 $ &
				$ \mathbf{1.5 \pm 1.0} $ &
				$ \mathit{2.2 \pm 1.4} $ &
				$ \mathbf{1.8 \pm 1.7} $ &
				$ \mathit{3.0 \pm 1.5} $ &
				$ 5.2 \pm 3.2 $ &
				$ \mathit{3.2 \pm 2.0} $ &
				$ \mathit{4.6 \pm 3.9} $ &
				$ \mathit{3.1 \pm 1.9}$ \\
				\hline												
				\multicolumn{11}{l}{\textit{$\star$ Standard deviations are not reported separately for the ``Expressions'' subset of the dataset.}} \\				
			\end{tabular}
		}
	\end{center}
\end{table*}

\begin{figure}[!t]
	\centering
	\includegraphics[width=0.99\linewidth]{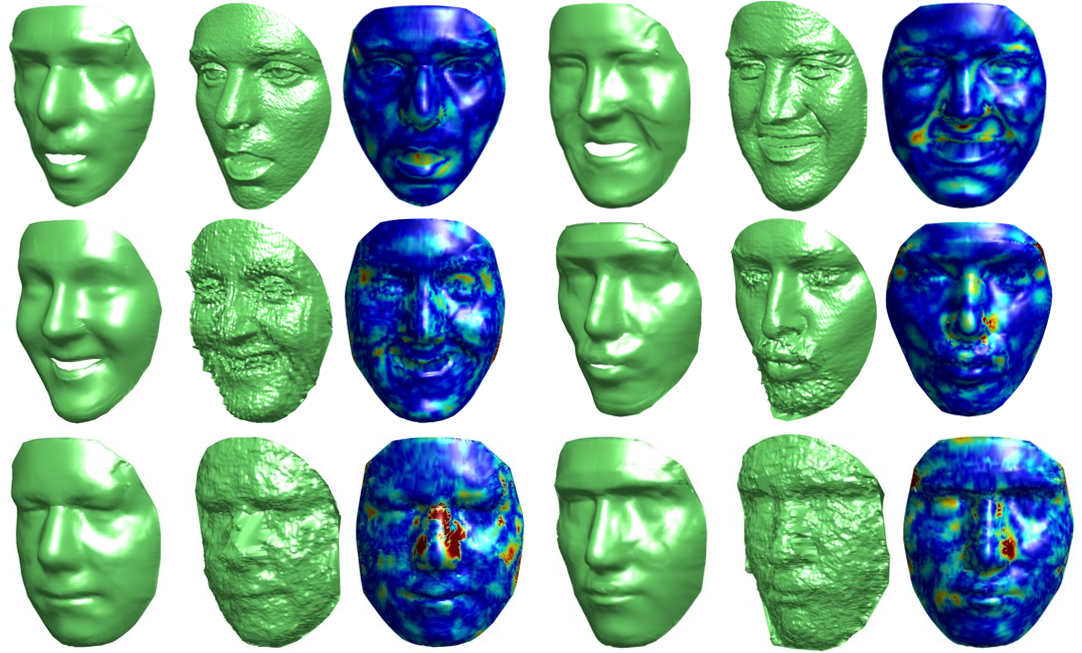}	
	\caption{\label{fig:defShapes} Fitted shapes $\mathbf{s}$ (first, fourth columns) and the corresponding ground-truth faces (second, fifth columns). FRGCv2.0 (first row), Bosphorus (middle row) and FaceWarehouse (bottom row). The error heatmaps show the distribution of the per-vertex error.}
\end{figure}

Plots in Figure~\ref{fig:cumlmErr} show that the proposed mean-point association and outliers rejection strategies, independently of the method used to learn the deformation components, lead to a more accurate landmark localization with respect to methods relying on a nearest-neighbor criterion, \ie,~NN+SLC and NICP. Consistently with the literature, NICP performs competitively on FRGCv2.0, but struggles when more complex expressions are considered, as in Bosphorus or FaceWarehouse. The CPD algorithm performs badly in all the cases. Results suggest the proposed mean-point association allows handling large shape differences and improving semantic alignment. Qualitative examples are depicted in Figure~\ref{fig:com_nn_mp}. Our solution demonstrated decisive for maintaining semantic consistency in these cases, especially in the mouth region. Some examples of fitted shapes $\mathbf{s}$ are instead reported in Figure~\ref{fig:defShapes}. Despite the presence of different noise sources (\eg,~beard, scanner noise) and large expressions, our algorithm results in smooth yet accurate fittings, without any landmarks based initialization.

Finally, we evaluate the NRF algorithm when used in conjunction with different 3DMM components. All the three variants attain pretty accurate detection, performing better than previous methods. When dealing with non-standard facial movements as those contained in FaceWarehouse though, the SLC-3DMM performs best, underlying the improved generalization to complex and unseen expressions, and resilience to high noise levels.

\subsubsection{Comparison with state-of-the-art}\label{subsec:sota-comp-landmark}
Though our approach was not specifically tailored for detecting landmarks, it is relevant to assess to what extent it is accurate in such task, as this represents a well-established indicator of the dense registration accuracy. To this end, we compare our approach with model-based state-of-the-art works in terms of landmarks detection on the FRGCv2.0 and Bosphorus. Results for the compared approaches have been collected from the original papers.

Table~\ref{tab:sota-landmarks-frgc} reports results on the FRGCv2.0. Overall, our approach can localize landmarks with sufficient precision, performing better than other model-based methods. Note that this represents a challenging cross-dataset scenario, as our 3DMM variants are built from the registered scans of the BU-3DFE dataset. In this regard, Gilani~\etal~\cite{Gilani-dense:2018} built their K3DMs from different datasets. We report results of the K3DM$_{BU}$, which is also built from the BU-3DFE for fair comparison.

\begin{table*}[t]
	\caption{\label{tab:sota-landmarks-warehouse} FaceWarehouse: Mean $\pm$ Std of landmark localization error (mm). \textbf{Ex/En}-outer/inner eye corner, \textbf{Prn}-nose-tip, \textbf{Sn}-nasal base, \textbf{Ac}-nose corner, \textbf{Ch}-mouth corner, \textbf{Ls/Li}-upper/lower lip. Left and right landmarks for Ex-En-Ch-Ac have been averaged. Best results in bold.}
	\begin{center}
		\scalebox{0.99}{
			\begin{tabular}{lcccccccccccc}
				\hline
				\textbf{Method} & \textbf{Ex} & \textbf{En} & \textbf{Prn} & \textbf{Sn} & \textbf{Ac} & \textbf{Ch} & \textbf{Ls} & \textbf{Li} & \textbf{Avg} \\
				\hline
				CPD~\cite{myronenko2010point} &
				$ 6.8 \pm 3.2 $ &
				$ 5.3 \pm 2.7 $ &
				$ 3.4 \pm 1.5 $ &
				$ 6.4 \pm 4.1 $ &
				$ 8.8 \pm 4.7 $ &
				$ 14.1 \pm 5.4 $ &
				$ 6.6 \pm 3.2 $ &
				$ 9.6 \pm 3.8 $ &
				$ 7.6 \pm 3.6 $ \\
				NICP ~\cite{Amberg-nicp:2007}&
				$ 5.4 \pm 3.0 $ &
				$ 3.4 \pm 2.1 $ &
				$ 3.0 \pm 1.9 $ &
				$ 3.6 \pm 2.8 $ &
				$ \mathbf{5.1 \pm 4.3} $ &
				$ 7.3 \pm 3.9 $ &
				$ 4.3 \pm 2.7 $ &
				$ 5.5 \pm 3.4 $ &
				$ 4.7 \pm 1.5 $ \\
				\hline
				NRF+PCA &
				$ 5.2 \pm 2.9 $ &
				$ 3.5 \pm 2.2 $ &
				$ 2.8 \pm 1.6 $ &
				$ 3.6 \pm 2.8 $ &
				$ 6.6 \pm 4.5 $ &
				$ 6.3 \pm 3.4 $ &
				$ 3.7 \pm 2.4 $ &
				$ 5.3 \pm 3.9 $ &
				$ 4.6 \pm 3.1 $ \\
				NRF+DL~\cite{ferrari:2015} &
				$ 5.1 \pm 2.8 $ &
				$ 3.0 \pm 1.9 $ &
				$ \textbf{2.6} \pm \textbf{1.6} $ &
				$ 3.5 \pm 2.7 $ &
				$ 6.1 \pm 4.5 $ &
				$ \textbf{5.8} \pm \textbf{3.2} $ &
				$ \textbf{3.4} \pm \textbf{2.4} $ &
				$ \mathbf{5.0 \pm 3.9} $ &
				$ 4.3 \pm 2.9 $ \\
				NRF+SLC &
				$ \textbf{4.0} \pm \textbf{2.4} $ &
				$ \textbf{2.7} \pm \textbf{1.7} $ &
				$ 2.8 \pm 1.7 $ &
				$ \textbf{3.4} \pm \textbf{2.6} $ &
				$ 5.5 \pm 4.5 $ &
				$ 5.9 \pm 3.5 $ &
				$ 3.5 \pm 2.4 $ &
				$ 5.6 \pm 4.6 $ &
				$ \textbf{4.1} \pm \textbf{2.8} $ \\
				\hline													
			\end{tabular}
		}
	\end{center}
\end{table*}

The advantage of our solution with respect to previous works comes out more evidently in Table~\ref{tab:sota-landmarks-bosph}, where results on Bosphorus are shown. For the compared approaches, we reported the results obtained using the ``Expression'' subset, consisting of $2,920$ scans.
The largest improvement is obtained on the landmarks of the mouth region, especially the lower lip that, in case of examples with strong expressions, it is the one with largest variation. It is also worth noting that the K3DM$_{BO}$~\cite{Gilani-dense:2018} model is built using scans of Bosphorus, whereas we perform a cross-database evaluation (results on this subset were not available for K3DM$_{BU}$).

\begin{figure}[!t]
	\centering
	\includegraphics[width=0.99\linewidth]{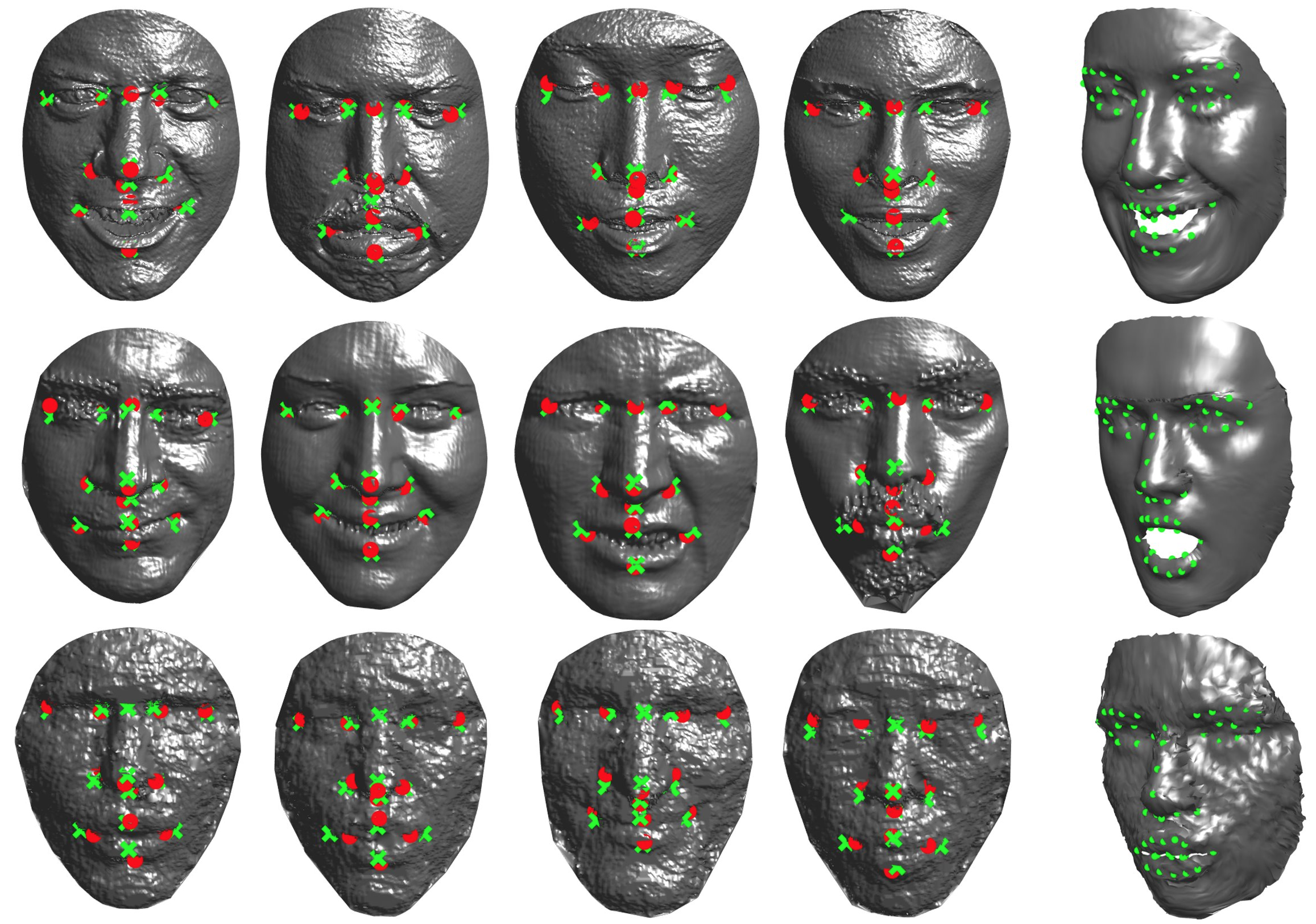}	
	\caption{\label{fig:lm_det_ex} Qualitative examples of landmark detection results on FRGCv2.0 (top row), Bosphorus (middle row), and FaceWarehouse (bottom row). Ground-truth annotations (red dots) and our transferred ones (green crosses) are plotted over the raw scans. Our approach is robust to different surfaces, noise, and topological changes. Some examples of re-indexed models $\mathbf{\hat{t}'}$ with an extended set of transferred landmarks are also shown (rightmost column).}
\end{figure}

We also report results on FaceWarehouse in Table~\ref{tab:sota-landmarks-warehouse}. We cannot compare with other methods as, to the best of our knowledge, we are the first to perform 3D landmark detection on this dataset. Results are computed with respect to the landmark annotations provided therein. In this dataset, point clouds are characterized by a high degree of noise, and show particularly complex expressions. Similar to Figure~\ref{fig:cumlmErr}, the higher generalization of SLC turns out here in a more evident way. Overall, NRF with all the three 3DMM solutions performs better than other approaches.

Qualitative examples of landmark annotation transfer can be appreciated in Figure~\ref{fig:lm_det_ex}. To highlight the capability of maintaining the semantic consistency in presence of large topological changes, in the rightmost column of this figure we show a few re-indexed models with an extended set of transferred landmarks.
Finally, we remark that all the reported detections for our method are obtained as a result of the dense registration process. Landmark detection is instead addressed independently in the compared works.

\subsection{Heterogeneous Large-scale Morphable Model}\label{subsec:3dmm-6K}
We have discussed that the more variability is contained in the training data, the more powerful and descriptive the morphable model.
The accuracy of the registration process results crucial for this purpose. Thus, we verify the validity of our dense registration approach by building a standard PCA model from the dense corresponded scans of FRGCv2.0, Bosphorus and FaceWarehouse, and evaluating its characteristics. To compare the different algorithms and demonstrate the value of our proposal, we build four different models using, respectively, our proposed NRF in conjunction with SLC, PCA, and DL~\cite{ferrari:2015} deformation components, and NICP, which is the standard solution employed in the vast majority of literature works~\cite{ploumpis2020towards, booth:2016}.
We considered all the available scans, for a total of $9,927$ training samples, comprising $721$ individuals with posed and spontaneous facial expressions, and a large variability in terms of age and ethnicity. To evaluate the intrinsic characteristics of this \textit{Heterogeneous Large Scale Morphable Model} (HLS-3DMM), we follow a common practice in the literature of statistical shape models and use \emph{compactness}, \emph{generalization} and \emph{specificity} measures~\cite{davies2008statistical}.
We underline here that performing a direct comparison with other large-scale models is not possible because of the very different topology of the templates. For example, the LSFM in~\cite{Booth:2017a} uses the BFM template~\cite{paysan20093d} that includes the ears and the eyeballs.
For the sake of completeness, we will comment and discuss the outcomes with respect to those works.

\begin{figure*}[!t]
	\centering
	\includegraphics[width=0.99\linewidth]{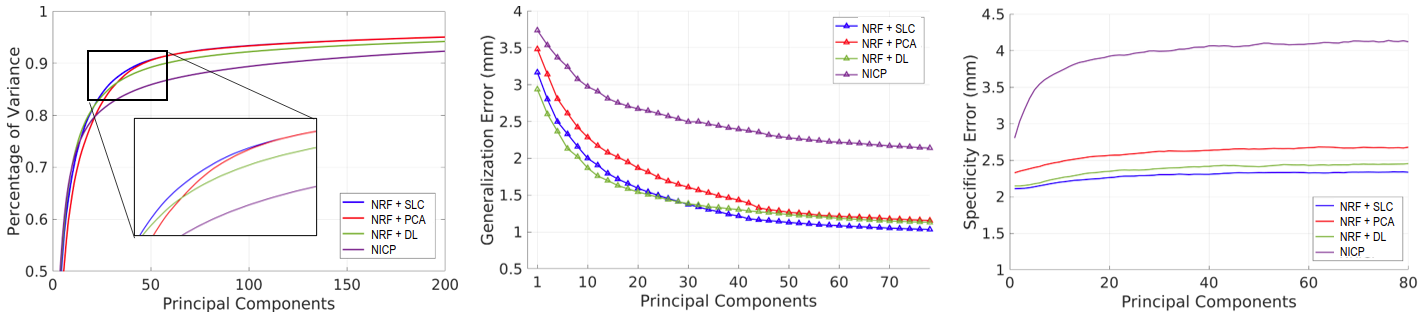}
	(a) Compactness
	\hspace{3.5cm}
	(b) Generalization
	\hspace{3.5cm}
	(c) Specificity	
	\caption{\label{fig:intrinsic} Compactness, generalization and specificity measures for the HLS-3DMM constructed from $9,927$ registered scans.}
\end{figure*}

\subsubsection{Compactness}\label{subsubsec:compactness}
\textit{Compactness} measures the percentage of variance of the training data that is explained by a model when a certain number of principal components are retained, and follows the principle ``\emph{the less, the better}''. It mainly depends on two factors: \emph{(i)} How precisely the dense registration algorithm maintained semantic consistency across the training samples; \emph{(ii)} The variance contained in the training data. In Figure~\ref{fig:intrinsic}~(a), we report the plots of the HLS-3DMM, comparing the model constructed from the scans registered with the proposed method and NICP.
The PCA bases derived using our registration algorithm result highly compact, with the SLC variant slightly outperforming PCA and DL. It explains more than 90\% of the variance with as few as 43 principal components (47 with PCA and 58 with DL). This result is consistent with~\cite{Booth:2017}, where the same variance is retained with approximately 40 components. Being the number of training samples similar ($\approx 10,000$), the data in~\cite{Booth:2017} contain about $9,000$ identities collected from the same device, but no expressions. Our training set has $471$ identities, but it includes much more variability in terms of posed and natural facial expressions. The most relevant outcome is that our models result much more compact with respect to NICP (90\% of the variance needs 114 components), indicating the significantly increased accuracy of registration.

\subsubsection{Generalization}\label{subsubsec:generalization}
\textit{Generalization} measures the ability of the model to represent novel instances of face shapes that are unseen during training. To this aim, following the standard practice, we split the registered scans into a train and a test set. The test set is composed of 300 scans, chosen randomly and in equal proportion from the three datasets, \ie,~100 from FRGCv2.0, 100 from Bosphorus and 100 from FaceWarehouse, to guarantee sufficient diversity.
The remaining identities are used for training so that identities between train and test do not overlap.
The different HLS-3DMMs are built applying PCA to the reduced training set. To compute the generalization error, for each shape in the test set, we estimate the deformation coefficients $\mathbf{\alpha}$ using~\eqref{eq:def-coefficients-estimation}, and deform the average model using~\eqref{eq:3dmmL}. The error is computed as the mean per-vertex Euclidean distance, and is shown in Figure~\ref{fig:intrinsic}~(b). We observe the capability of HLS-3DMM built using our solution to generalize well to unseen samples. The SLC variant performs better than the others from 30 components onward, suggesting more and variegated deformations were captured in the registration process. Other results in the recent literature report similar outcomes, with the UHM model in~\cite{ploumpis2020towards} having a generalization error ranging from 3 to 0.5$mm$, and the LSFM in~\cite{Booth:2017a} ranging from 1.8 to 0.3$mm$ for neutral scans only. Again, we obtain a significant improvement with respect to a model resulting from NICP.

\subsubsection{Specificity}\label{subsubsec:specificity}
\textit{Specificity} evaluates the validity of synthetic faces generated by the model. We randomly generated $1,000$ synthetic faces for a fixed number of principal components and measured how close they are to the samples in the test set. For each synthetic face, we found the sample in the test set with minimum error in terms of (average) per-vertex Euclidean distance. The curve in Figure~\ref{fig:intrinsic}~(c) reports the average of this error across all the synthetic faces. In comparison with other works, we attain good specificity error. For example, Booth~\etal~\cite{Booth:2017a} reported errors varying from 1 to 1.8$mm$ on a test set of neutral models only. Ploumpis~\etal~\cite{ploumpis2020towards}, instead, reported errors varying from 2.25 to 3.25$mm$ for their UHM. Our model is capable of generating realistic faces, which also include facial expressions. As a desirable behavior, we observe the error remains stable even for a larger number of components. Our three models, again, outperform NICP significantly.

\subsection{Comparing with Previous Models}\label{subsec:hls_vs_bufe}
\begin{figure}[!t]
	\centering
	\includegraphics[width=0.8\linewidth]{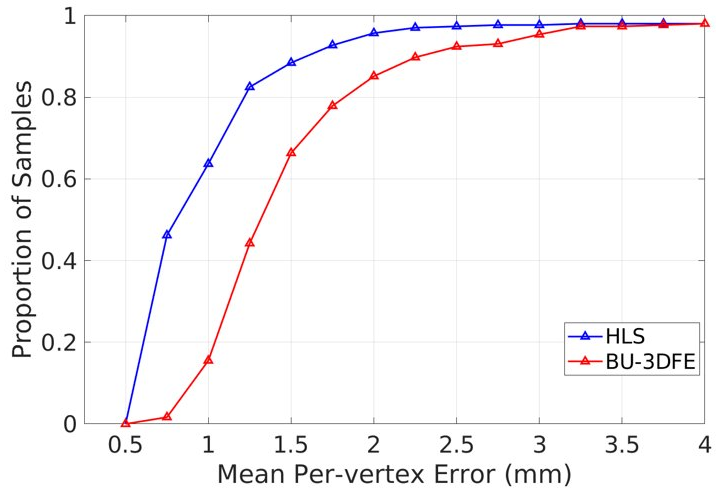}	
	\caption{\label{fig:fitt_app_bufe} Cumulative error distribution of the per-vertex fitting error between PCA models built from the newly registered $9,927$ scans (HLS), and $1,779$ scans of BU-3DFE.}
\end{figure}

As last evaluation, we showcase the increased modeling capability of the HLS-3DMM against a PCA model constructed from the $1,779$ registered scans of BU-3DFE as obtained in~\cite{ferrari:2015}. To this aim, we use the same 300 scans composing the test set defined in Section~\ref{subsubsec:generalization}, and compare the fitting accuracy of the two models. Without loss of generality, we perform the fitting and compute the error in the same way as in Section~\ref{subsubsec:generalization}\footnote{The comparison is possible as all the scans share the same topology.}. For both, the number of principal components retaining the 99\% of the variance are used. Figure~\ref{fig:fitt_app_bufe} clearly shows a significantly improved accuracy with respect to the BU-3DFE model. This is attributed both to the larger training set used, and the accuracy of the dense registration between scans. The significantly larger variability included in the model allows it to generalize well to a wider variety of faces, even though the BU-3DFE dataset already contained posed expressions in the training set.

\section{Conclusions}\label{sec:conclusion}
In this paper, we proposed a dense 3D face correspondence approach, which leverages a 3DMM to transfer a point-to-point semantic annotation across raw 3D faces. We proposed a novel formulation to learn sparse and locally coherent deformation components that showed noteworthy generalization potential to new and unseen facial deformations. We showed that, by radically changing the interpretation of the training data and treating each vertex in the scans as an independent sample, it is possible to learn sparse and decorrelated facial motions with high generalization potential.
The model deformation is guided by the proposed Non-Rigid Fitting algorithm, specifically tailored for adapting to heterogeneous scans and large shape differences. The approach demonstrated to be accurate even in the presence of strong facial expressions, and for very different types of 3D faces without the need of any landmark based guidance. We then densely registered together scans coming from three databases, and built a large-scale, heterogeneous 3DMM from them, with promising modeling capabilities. The very different and complementary characteristics of the considered databases suggest the approach can be used to further enlarge the set of registered scans to, potentially, any database already collected.

\ifCLASSOPTIONcaptionsoff
  \newpage
\fi


\bibliographystyle{IEEEtran}
\bibliography{bibliography}
%

%

\begin{IEEEbiography}[{\includegraphics[width=1in,height=1.25in,clip,keepaspectratio]{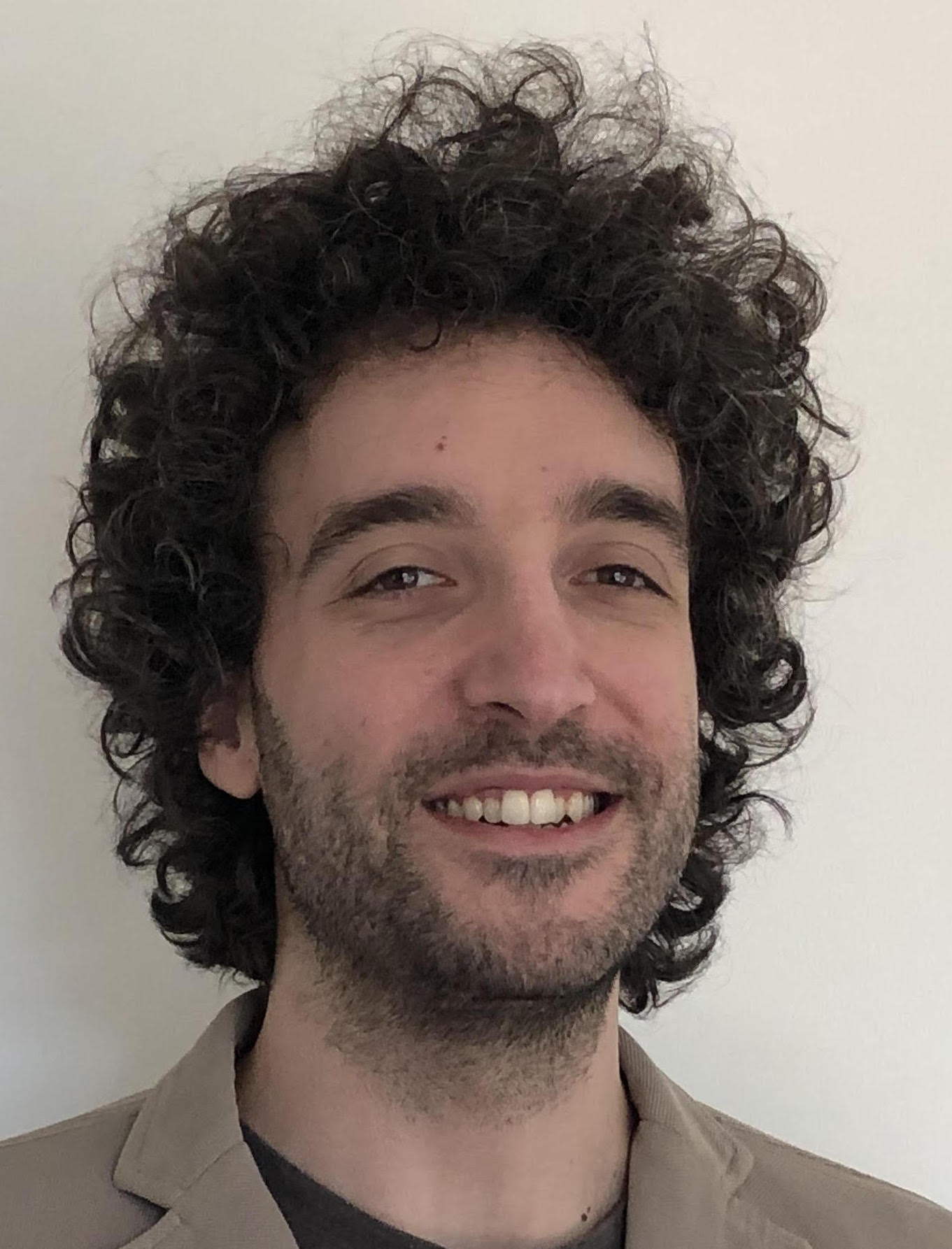}}]{Claudio Ferrari} received the Ph.D. in Information Engineering from the University of Florence in 2018. Currently, he is a postdoctoral researcher at the Media Integration and Communication Center (MICC) of the University of Florence. He has been a visiting research scholar at the University of Southern California (USC) in 2014. His research interests focus on computer vision and machine learning for biometrics and 2D/3D face analysis.
\end{IEEEbiography}

\begin{IEEEbiography}[{\includegraphics[width=1in,height=1.25in,clip,keepaspectratio]{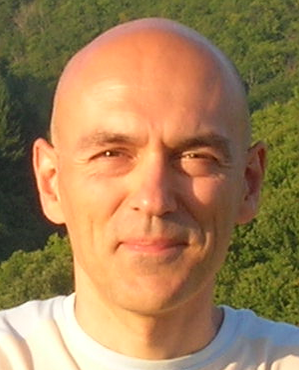}}]{Stefano Berretti} received the Ph.D. in Computer Engineering in 2001. Currently, he is an Associate Professor at University of Florence, Italy. He has been Visiting Professor at University of Lille and University of Alberta. His research interests focus on computer vision for face biometrics, human emotion and behavior understanding, computer graphics and multimedia. He is the Information Director and an Associate Editor of the ACM Transactions on Multimedia Computing, Communications, and Applications, and Associate Editor of the IET Computer Vision journal.
\end{IEEEbiography}

\begin{IEEEbiography}[{\includegraphics[width=1in,height=1.25in,clip,keepaspectratio]{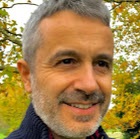}}]{Pietro Pala} received the Ph.D. in Information and Telecommunications Engineering in 1997 at the University of Florence. Currently, he is Full Professor of Informatics Engineering at the University of Florence. His research activity focus on the use of pattern recognition for multimedia information retrieval and biometrics, and on the study of 3D data for person and action recognition. He serves as editor for Multimedia Systems and ACM Transactions on Multimedia Computing, Communications, and Applications (TOMM).
\end{IEEEbiography}

\begin{IEEEbiography}[{\includegraphics[width=1in,height=1.25in,clip,keepaspectratio]{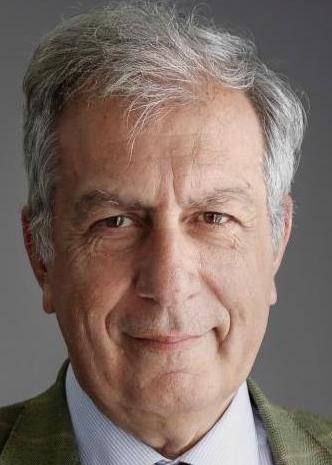}}]{Alberto Del Bimbo} is Full Professor of Computer Engineering at the University of Florence, Italy. His scientific interests include multimedia retrieval, pattern recognition, image and video analysis and human–computer interaction. Prof. Del Bimbo is IAPR Fellow, Associate Editor of several journals in the area of pattern recognition and multimedia, and the Editor-in-Chief of the ACM Transactions on Multimedia Computing, Communications, and Applications. He was also the recipient of the prestigious SIGMM 2016 Award for Outstanding Technical Contributions to Multimedia Computing, Communications and Applications.
\end{IEEEbiography}




\end{document}


\title{Supplemental Material to the paper: \\ A Sparse and Locally Coherent Morphable Face Model for Dense Semantic Correspondence Across Heterogeneous 3D Faces}


\author{Claudio Ferrari,~\IEEEmembership{Member,~IEEE,}
        Stefano Berretti,~\IEEEmembership{Senior,~IEEE,}
        Pietro Pala,~\IEEEmembership{Senior,~IEEE,} \\
        and~Alberto Del~Bimbo,~\IEEEmembership{Senior,~IEEE}
\IEEEcompsocitemizethanks{\IEEEcompsocthanksitem C. Ferrari, S. Berretti, P. Pala and A. Del~Bimbo are with the Department of Information Engineering, University of Florence, Florence, Italy, 50139.\protect\\
E-mail: claudio.ferrari@unifi.it}
\thanks{Manuscript received ; revised.}}

%
%

\markboth{IEEE Transactions on Pattern Analysis and Machine Intelligence,~Vol.~??, No.~??, May~2020}%
{C. Ferrari \MakeLowercase{\etal}: Supplemental Material: Building a Heterogeneous, Large Scale Morphable Face Model}
\maketitle

\IEEEdisplaynontitleabstractindextext

%
\IEEEpeerreviewmaketitle

\section{SLC Parameters}\label{subsubsec:slc-params}
\begin{figure*}[!t]
	\centering
	\includegraphics[width=0.99\linewidth]{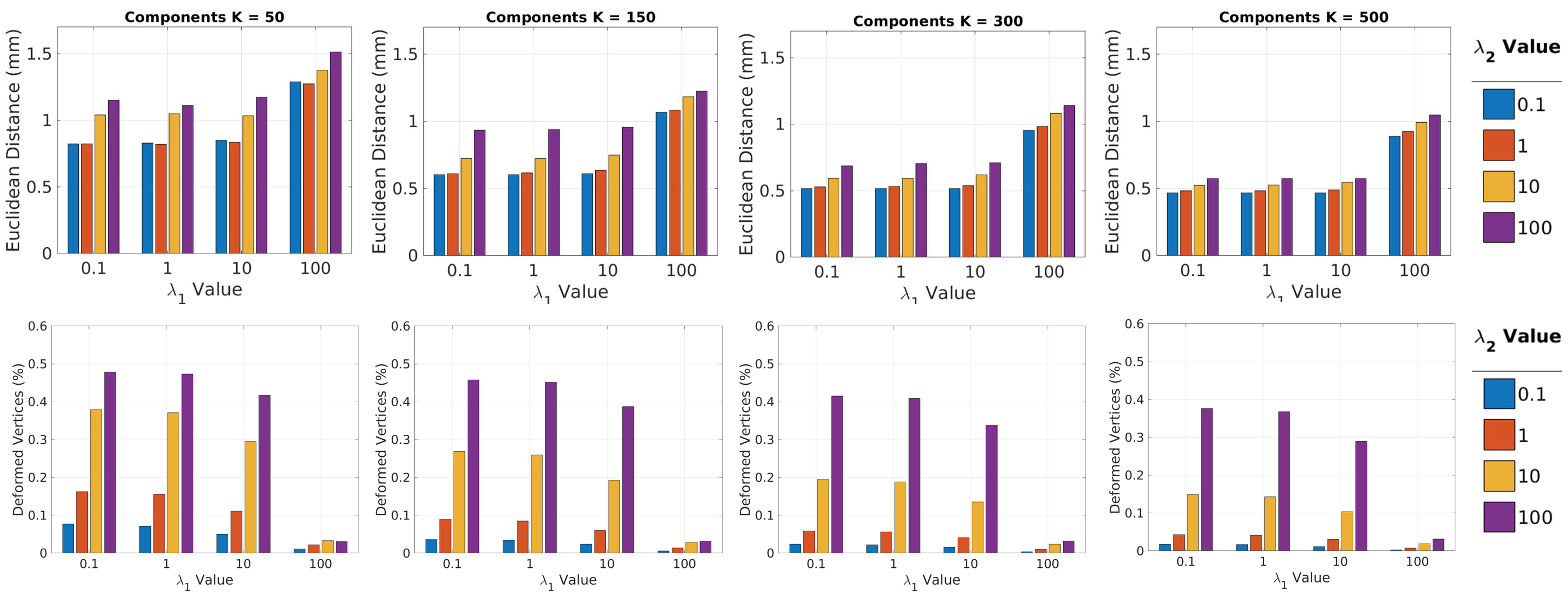}
	\caption{\label{fig:ablation} Sensitivity analysis on the SLC parameters. Plots report the per-vertex error obtained on a sample of $200$ scans from the FRGCv2.0 database applying NRF (top row) with SLC components learned varying $\lambda_1$ and $\lambda_2$. In the bottom row, the average proportion of deformed vertices with respect to $\lambda_1$ and $\lambda_2$ are reported. Each plot reports a different $k$.}
\end{figure*}

In this section, we report an additional ablation study on the SLC parameters. Parameters $\lambda_1$ and $\lambda_2$ mainly define the spatial extent of each SLC component. The best choice of such parameters follows two observations: (\emph{i}) Each component needs to be sufficiently limited so that facial regions are deformed independently. This also maintains the founding concept of local consistency of motion; (\emph{ii}) We still want the spatial extent to be large enough to cover the entire shape yet without introducing irregularities. Here, we explore the effect of learning parameters in a real fitting scenario. In Figure~\ref{fig:ablation}~(top row), we report the reconstruction error obtained by applying NRF to a subset of $200$ scans of the FRGCv2.0, using components learned with different values for $\lambda_1$, $\lambda_2$ and $k$. The error is computed as the average per-vertex, nearest neighbor Euclidean distance between the fitted model $\mathbf{s}$ and $\mathbf{\hat{t}}$. In this test, all the $1,779$ training scans were used to build the SLC-3DMM. First, we note the per-vertex error is generally low and inversely proportional to $k$. Using all the available training scans, even with as few as $50$ components, the reconstruction error is yet reasonably low (less than 1mm). Parameters $\lambda_1$ and $\lambda_2$, instead, behave consistently across different $k$. For each $k$, the highest per-vertex error is obtained when $\lambda_2$ is very large ($\lambda_2 = 100$), confirming that each component should be spatially bounded. The error decreases when the two parameters are balanced and kept stable to lower values ($<10$), so that reasonably small regions are smoothly deformed. In fact, excessively large deformations ($\lambda_2$) imply undesired interactions, whereas smaller, non-smooth deformations ($\lambda_1$) can prevent the model capturing global shape variations. Despite more components increase the modeling diversity, which can be helpful to obtain accurate fittings, the non-negligible presence of noise in raw scans can impact on the smoothness of the result. Since our main goal is to densely register together a large number of heterogeneous scans, we privileged a configuration ensuring higher smoothness. So, we employed a SLC configuration with $k=50$, $\lambda_1=10$, and $\lambda_2=1$ for the experiments reported in the main document.

\begin{figure}[!t]
	\centering
	\includegraphics[width=0.97\linewidth]{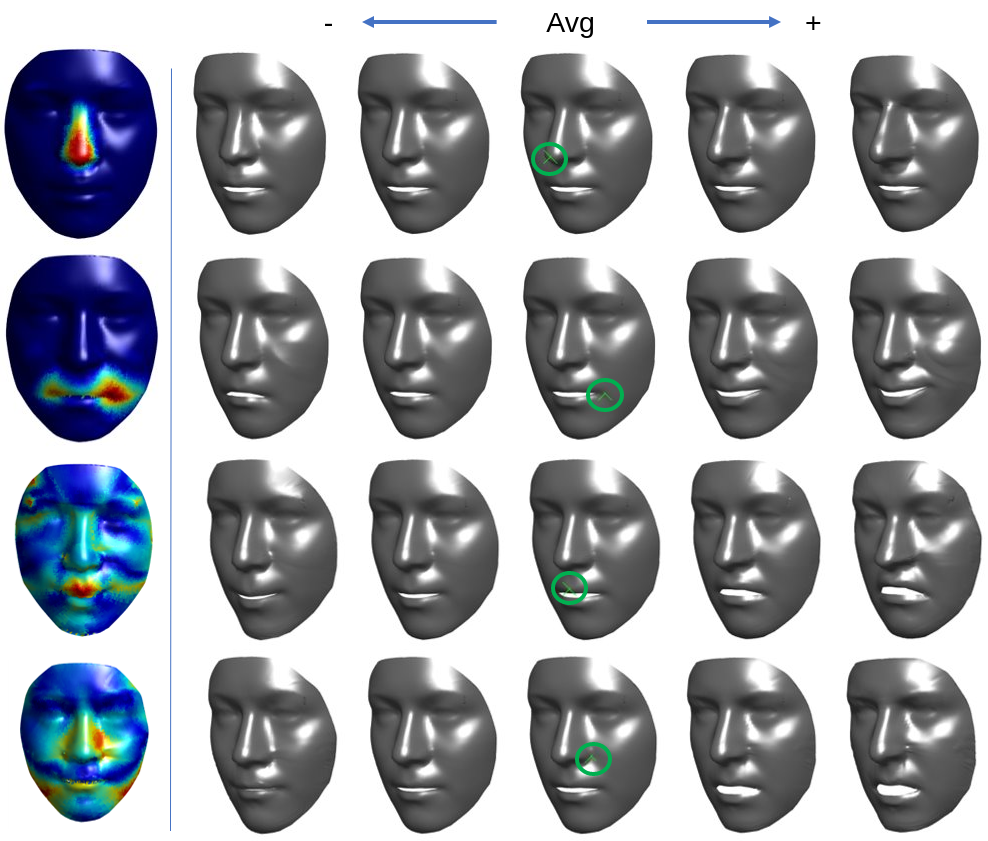}	
	\caption{\label{fig:control_def} The average model can be locally deformed by selecting one or more control points. SLC components (top two rows) affect the region surrounding the control point (green circle). The PCA model (third row) and the DL-3DMM~\cite{Ferrari:2017b} instead affect the whole shape.}
\end{figure}

Figure~\ref{fig:ablation}~(bottom row) instead reports a quantitative evaluation of the extent of components deformation as a function of $\lambda_1$, $\lambda_2$ and $k$. A valuable property of our solution is that by increasing $k$, the number of vertices that each component affects diminishes. This effect has a major benefit: it allows us to increase the number of components $k$ and model finer yet complementary surface areas. Ultimately, this prevents the performance to saturate, and significantly increases the modeling generalization capabilities.

\section{Controlling the SLC Components}\label{subsubsec:slc_visualize}
In the following, we show that the local and anatomically meaningful nature of the SLC components can be exploited to manipulate a 3D face in an intuitive way. Each component $\mathbf{C}_i$ applies a deformation to a spatially restricted region. Thus, we can select which component to use depending on the area of the face we wish to warp. To show this, we choose one control vertex $m_c$ in the average model, select the subset of $k^\star$ components ($k^\star < k$) that are not null in $m_c$, and deform $\mathbf{m}$ with a single or a combination of some of the selected $k^\star$ components. For the sake of clearness, in this application we set $k=300$.

Considering 14 landmarks corresponding to significant areas of the face~\cite{Creusot-ijcv:2013}, the average number of active SLC components in correspondence of each one is $k^\star = 86$ out of $k=300$.
Figure~\ref{fig:control_def} shows a qualitative example on some of these landmarks. Note the SLC components (top two rows) affect only the region surrounding the control point yet resulting in natural looking deformations. The standard PCA model (third row) and the DL-3DMM~\cite{Ferrari:2017b}, instead, deform the whole shape. This is a valuable property of our model that increases its versatility also in different application scenarios. It is also worth to note the the SLC components can effectively decouple and reproduce structural deformations related to identity traits (nose shape in Figure~\ref{fig:control_def}, top row) as well as expressions (asymmetric smile in Figure~\ref{fig:control_def}, second row).

\section{Handling Occlusions and Rotations}\label{subsec:occ_rot}
Some previous works in the literature~\cite{Creusot-ijcv:2013, sukno20143, Gilani-dense:2018} reported landmark detection results for the occluded and rotated scans in the Bosphorus dataset. An accurate detection of landmarks in this particular scenario is generally achieved by excluding missing or occluded regions from the fitting process, as done for example by Gilani~\etal~\cite{Gilani-dense:2018}, or using local surface properties like normals orientation or local curvature features~\cite{Creusot-ijcv:2013}. However, this works well as long as the final objective is the detection of a sparse set of landmarks, or the reconstruction of an approximate shape of the face. We decided not to include such discussion and results in the main manuscript since our approach has been specifically designed to put in dense correspondence scans from diverse databases, so accounting for the heterogeneity in their structures. In this context, including occluded or rotated scans (with missing face regions) does not make much sense and the definition of dense correspondence becomes ambiguous. In fact, other works addressing the dense correspondence with the aim of collecting large training datasets for building a 3DMM~\cite{Booth:2017a, ploumpis2019combining} usually filter out possibly wrong examples to make sure the training data is as clean as possible.

\begin{table*}[t]
	\caption{\label{tab:sota-landmarks-bosph} Mean and standard deviation of landmark localization error (mm) on the Bosphorus dataset. \textbf{Ex/En}-outer/inner eye corner, \textbf{N}-nose bridge, \textbf{Prn}-nose-tip, \textbf{Sn}-nasal base, \textbf{Ac}-nose corner, \textbf{Ch}-mouth corner, \textbf{Ls/Li}-upper/lower lip. Landmarks for Ex-En-Ch-Ac have been averaged. Best results in bold}
	\begin{center}
		\scalebox{0.85}{
			\begin{tabular}{llccccccccccc}
				\hline
				& \textbf{Method}  & \textbf{Ex} & \textbf{En} & \textbf{N} & \textbf{Prn} & \textbf{Sn} & \textbf{Ac} & \textbf{Ch} & \textbf{Ls} & \textbf{Li} & \textbf{Avg} \\
				\hline
				\multirow{4}{*}{Occlusions}
				& Cruesot \textit{et al.}~\cite{Creusot-ijcv:2013}  &
				$ 6.7 $ &
				$ 5.3 $ &
				$ 7.8 $ &
				$ 4.7 $ &
				$ 11.0 $ &
				$ 5.1 $ &
				$ 4.8 $ &
				$ 4.8 $ &
				$ 4.3 $ &
				$ 6.0 $ \\
				& Sukno \textit{et al.}~\cite{sukno20143}  &
				$ 6.4 $ &
				$ 3.8 $ &
				$ 4.1 $ &
				$ 3.8 $ &
				$ 3.8 $ &
				$ 4.5 $ &
				$ 4.9 $ &
				$ 3.6 $ &
				$ 4.8 $ &
				$ 4.4 $ \\
				& K3DM$_{BO}$~\cite{Gilani-dense:2018} &
				$ \mathbf{4.6} $ &
				$ \mathbf{3.0} $ &
				$ \mathbf{2.7} $ &
				$ \mathbf{3.2} $ &
				$ 2.9 $ &
				$ \mathbf{2.8} $ &
				$ \mathbf{4.3} $ &
				$ 2.9 $ &
				$ 4.1 $ &
				$ \mathbf{3.4} $ \\
				& \textbf{Ours+SLC}  &
				$ 5.8 \pm 4.4 $ &
				$ 5.1 \pm 3.1 $ &
				$ 3.7 \pm 2.3 $ &
				$ 4.3 \pm 2.0 $ &
				$ \mathbf{2.2 \pm 1.8} $ &
				$ 6.1 \pm 3.5 $ &
				$ 4.6 \pm 4.2 $ &
				$ \mathbf{2.6 \pm 2.2} $ &
				$ \mathbf{3.3 \pm 2.8} $ &
				$ 4.2 \pm 1.3$ \\
				\hline
				\multirow{4}{*}{Rotations}
				& Cruesot \textit{et al.}~\cite{Creusot-ijcv:2013}  &
				$ \mathbf{4.7} $ &
				$ 4.4 $ &
				$ 5.2 $ &
				$ 4.9 $ &
				$ 9.5 $ &
				$ 3.4 $ &
				$ 4.2 $ &
				$ 3.8 $ &
				$ \mathbf{3.8} $ &
				$ 4.9 $ \\
				& Sukno \textit{et al.}~\cite{sukno20143}  &
				$ \mathbf{4.7} $ &
				$ \mathbf{3.1} $ &
				$ 3.4 $ &
				$ 4.4 $ &
				$ 4.2 $ &
				$ 3.4 $ &
				$ \mathbf{3.7} $ &
				$ \mathbf{3.5} $ &
				$ 5.0 $ &
				$ 3.9 $ \\
				& K3DM$_{BO}$~\cite{Gilani-dense:2018} &
				$ 4.9 $ &
				$ 3.5 $ &
				$ \mathbf{2.7} $ &
				$ 3.2 $ &
				$ 3.8 $ &
				$ \mathbf{2.9} $ &
				$ 4.7 $ &
				$ 4.1 $ &
				$ 5.8 $ &
				$ \mathbf{4.0} $ \\
				& \textbf{Ours+SLC}  &
				$ 6.4 \pm 4.4 $ &
				$ 5.4 \pm 3.5 $ &
				$ 4.4 \pm 2.7 $ &
				$ \mathbf{2.4 \pm 1.3} $ &
				$ \mathbf{3.7 \pm 2.0} $ &
				$ 4.3 \pm 3.3 $ &
				$ 6.1 \pm 4.3 $ &
				$ \mathbf{3.5 \pm 1.8} $ &
				$ 4.8 \pm 3.2 $ &
				$ 4.5 \pm 1.3$ \\
				\hline		
			\end{tabular}
		}
	\end{center}
\end{table*}

\begin{figure}[!t]
	\centering
	\includegraphics[width=0.99\linewidth]{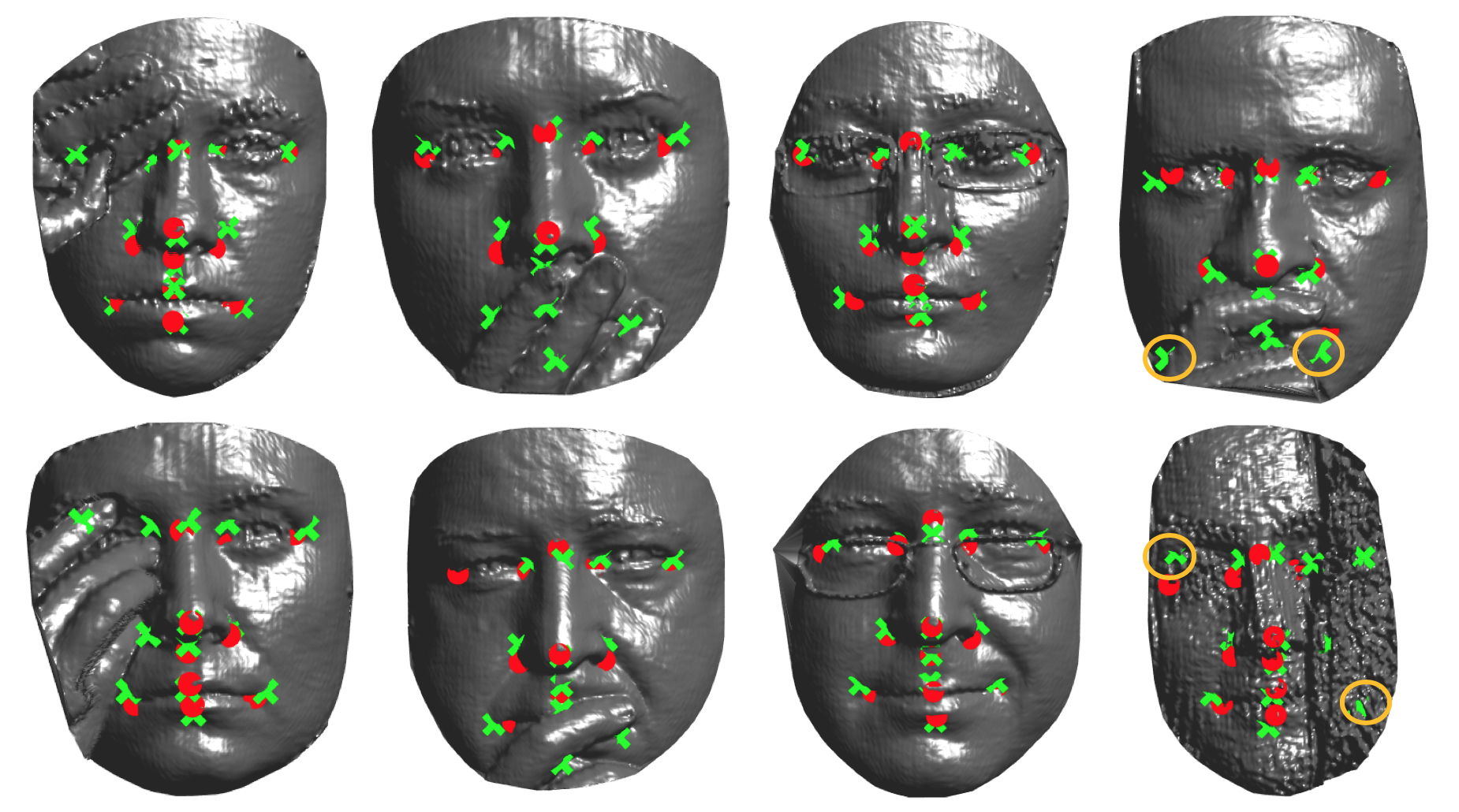}	
	\caption{\label{fig:lmDet_occ} Some examples of landmark detection on occluded scans. Green crosses: our result; red dots: ground-truth annotations. The visible landmarks can be detected with sufficient accuracy. On the other hand, the location of non visible landmarks is still predicted. Two partial failure cases are also shown (rightmost column).}
\end{figure}

\begin{figure}[!t]
	\centering
	\includegraphics[width=0.92\linewidth]{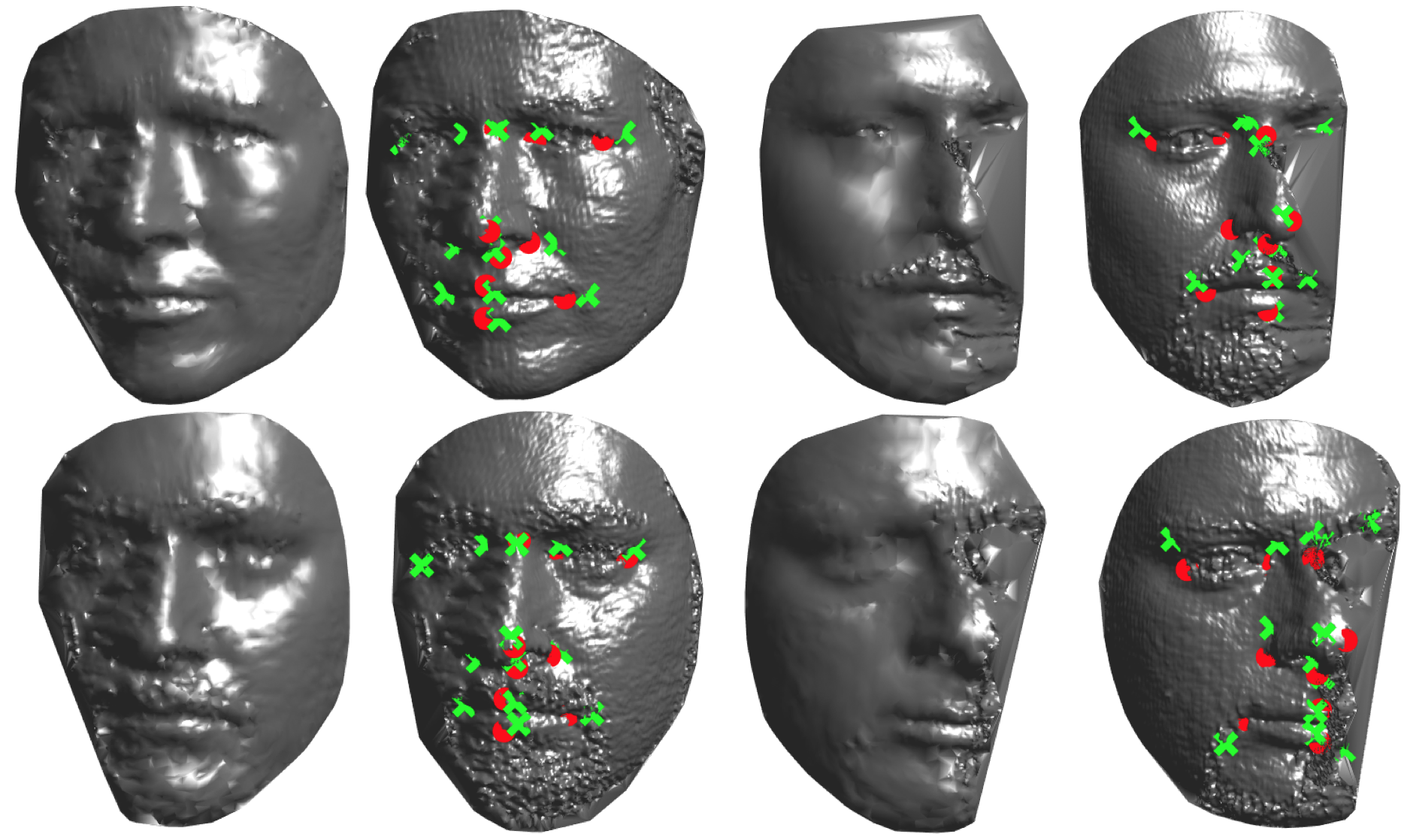}	
	\caption{\label{fig:lmDet_rot} Some examples of landmark detection on rotated scans. Green crosses: our result; red dots: ground-truth annotations. The re-indexed models $\mathbf{\hat{t}'}$ are also reported and show that the visible part of the face is registered fairly accurately.}
\end{figure}

For the sake of completeness, we report here both a quantitative and qualitative evaluation of landmark detection on the rotated and occluded scans of the Bosphorus dataset. The error is computed considering the available annotations of non-occluded landmarks. Quantitative results in comparison with the same methods as reported in the main document are shown in Table~\ref{tab:sota-landmarks-bosph}. From the results, it emerges that other approaches specifically tailored for detecting sparse landmarks can, on average, better localize them in case of occlusions or missing face regions due to self-occlusion. Still, our approach performs competitively, obtaining more accurate detections for some landmarks and a sufficient general level of accuracy. It is worth to note that the subset of landmarks obtaining worst localizations with respect to the other approaches are those associated to the corners of the eyes and the mouth (Ex, En, Ac). This because our method purposely attempts to keep the landmark configuration consistent within semantically equivalent areas. Whereas natural asymmetries are captured, in case of occlusions or large missing regions, the localization of the visible landmarks are partially compromised by the occlusions. This is qualitatively shown in Figure~\ref{fig:lmDet_occ}. A valuable feature that instead is not explicitly exposed by the other approaches is the capability of predicting the location of the occluded landmarks. Clearly, this is an ill-posed problem; nevertheless, we can qualitatively see that the locations of the predicted occluded landmarks, which do not have a corresponding manual annotation, are yet rather accurate. Figure~\ref{fig:lmDet_rot} shows qualitative examples on rotated scans. In this scenario, the same problems arise. However, the re-indexed models $\mathbf{\hat{t}'}$ look natural on the visible side, indicating the fitting process was not compromised, and a certain degree of robustness is achieved. On the other hand, the final semantic annotation transfer makes the vertices corresponding to self-occluded areas to collapse at the boundary, underlying the ambiguity of this scenario.

\section{HLS-3DMM}
\begin{figure*}[!t]
	\centering
	\includegraphics[width=0.8\linewidth]{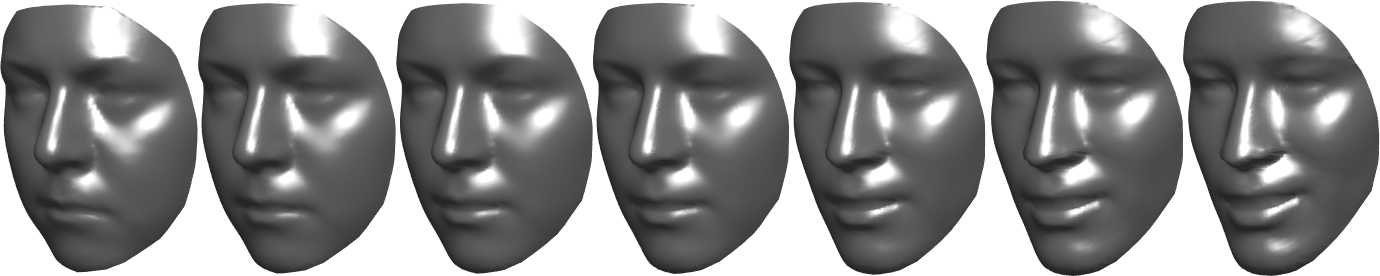}
	\includegraphics[width=0.8\linewidth]{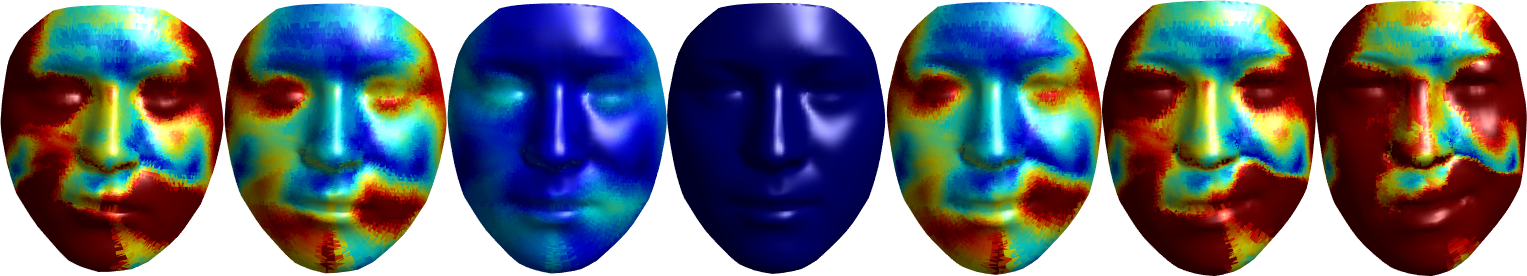}
	\includegraphics[width=0.8\linewidth]{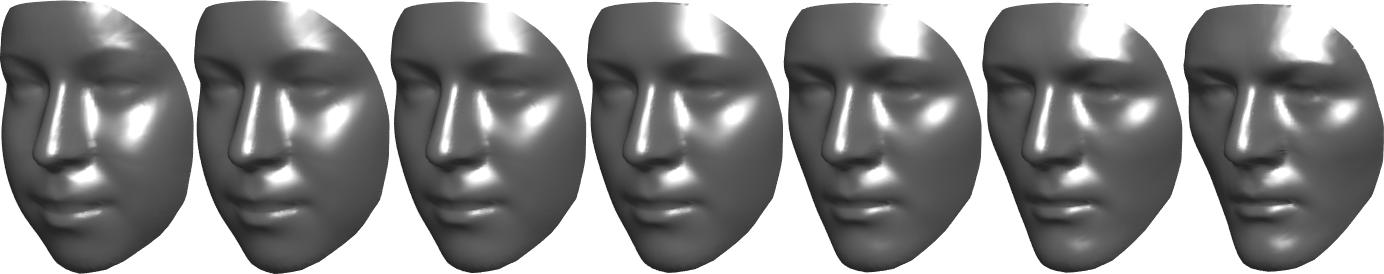}
	\includegraphics[width=0.8\linewidth]{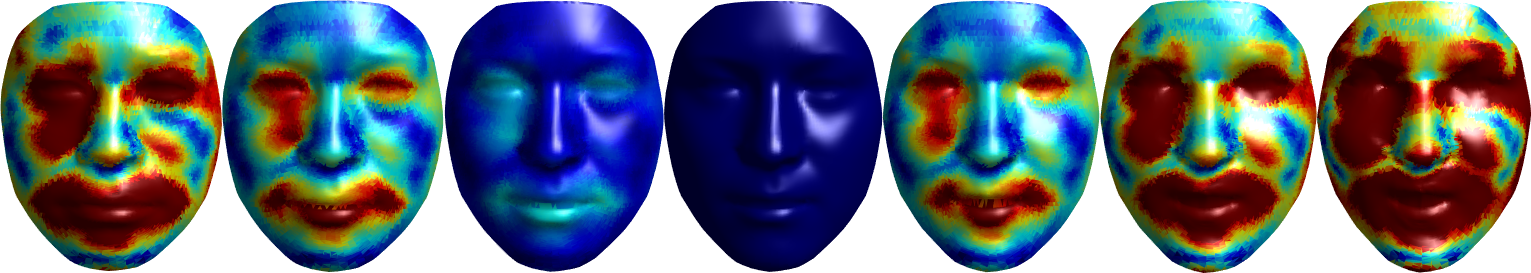}
	\includegraphics[width=0.8\linewidth]{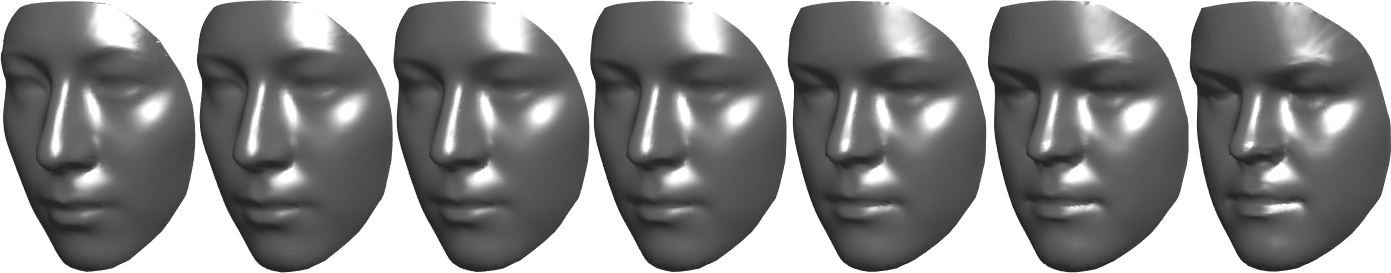}
	\includegraphics[width=0.8\linewidth]{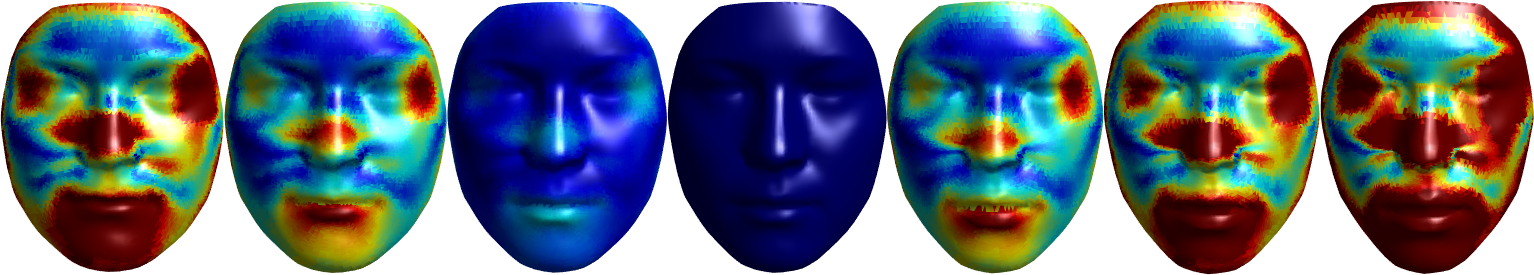}
	\includegraphics[width=0.8\linewidth]{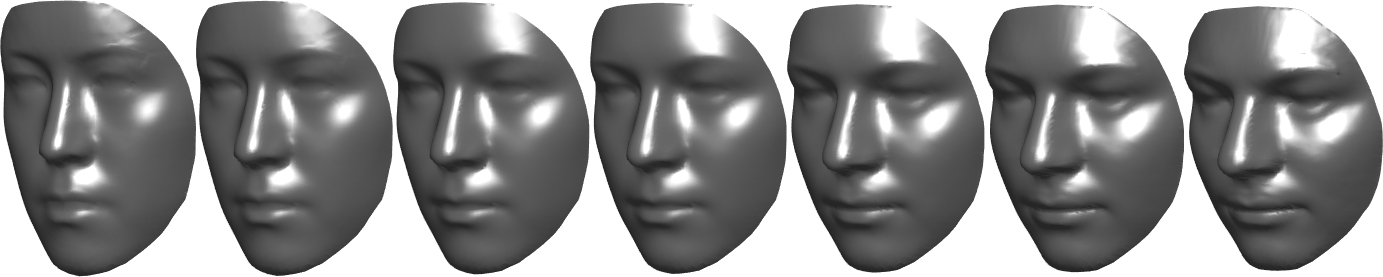}
	\includegraphics[width=0.8\linewidth]{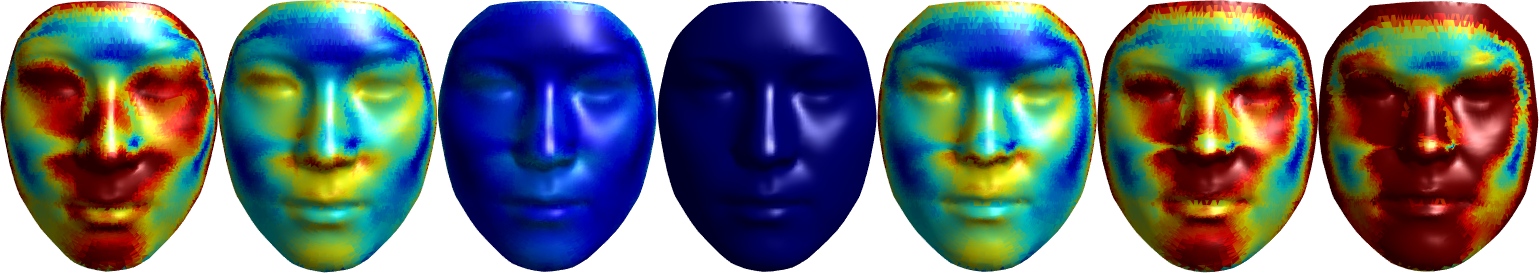}			
	\caption{\label{fig:3dmm_ex_pca} 3D shapes generated deforming the average model (middle column) by varying the top 4 PCA deformation components.}
\end{figure*}

\begin{figure*}[!t]
	\centering
	\includegraphics[width=0.8\linewidth]{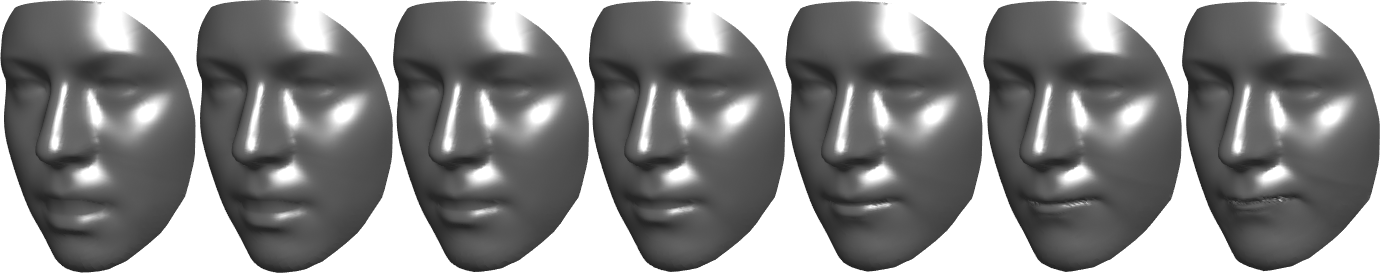}
	\includegraphics[width=0.8\linewidth]{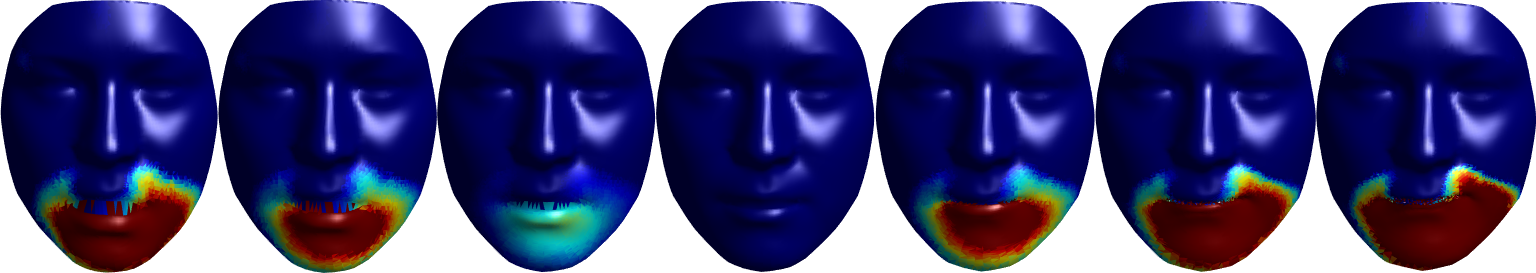}
	\includegraphics[width=0.8\linewidth]{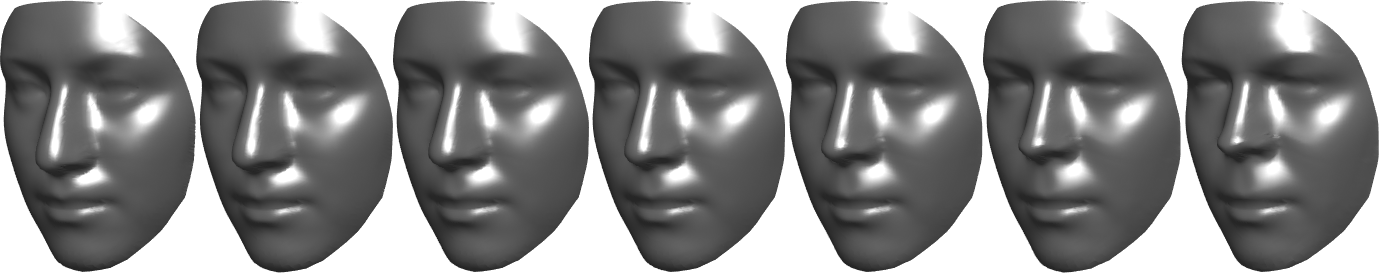}
	\includegraphics[width=0.8\linewidth]{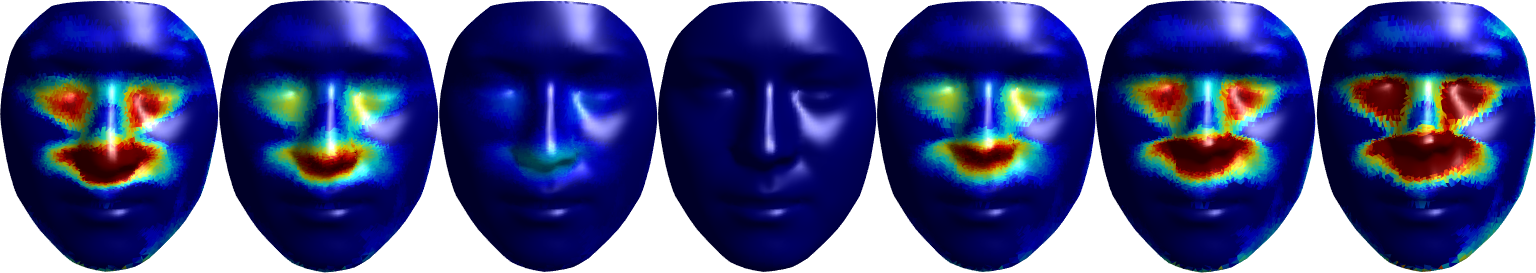}	
	\includegraphics[width=0.8\linewidth]{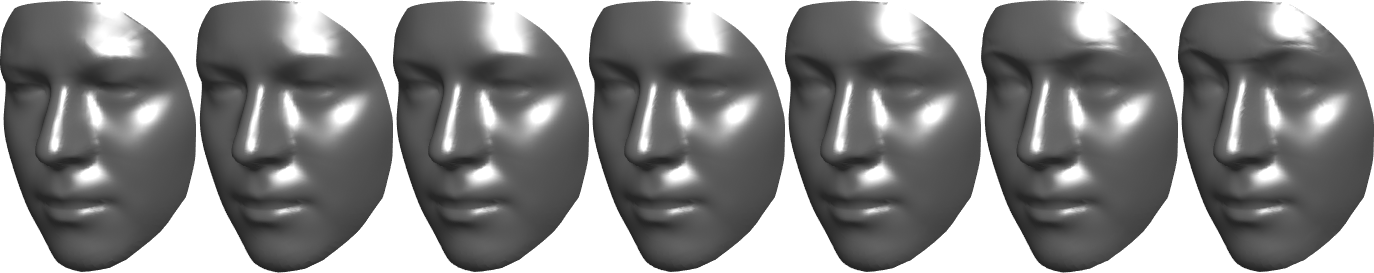}
	\includegraphics[width=0.8\linewidth]{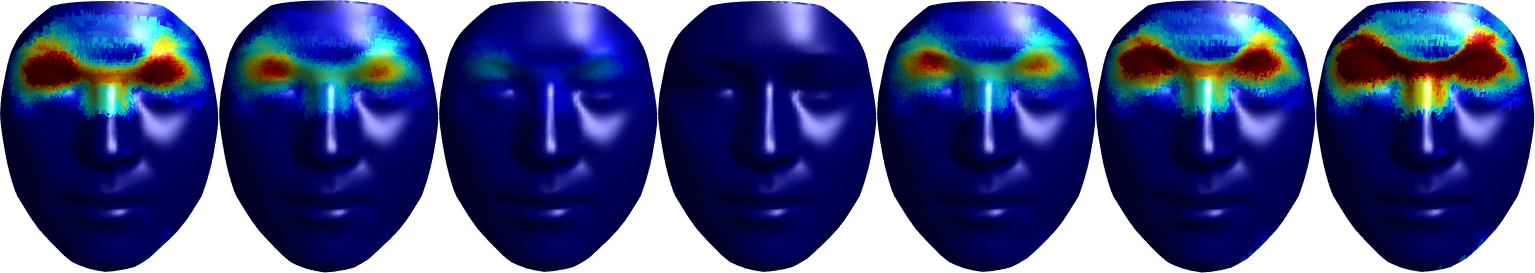}
	\includegraphics[width=0.8\linewidth]{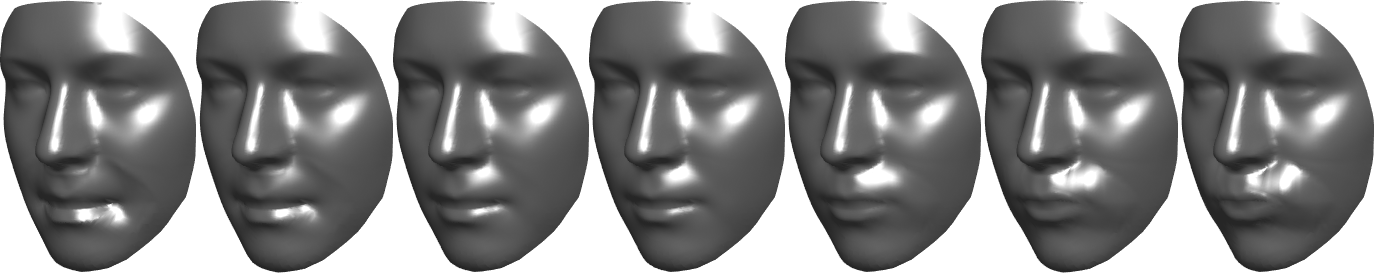}
	\includegraphics[width=0.8\linewidth]{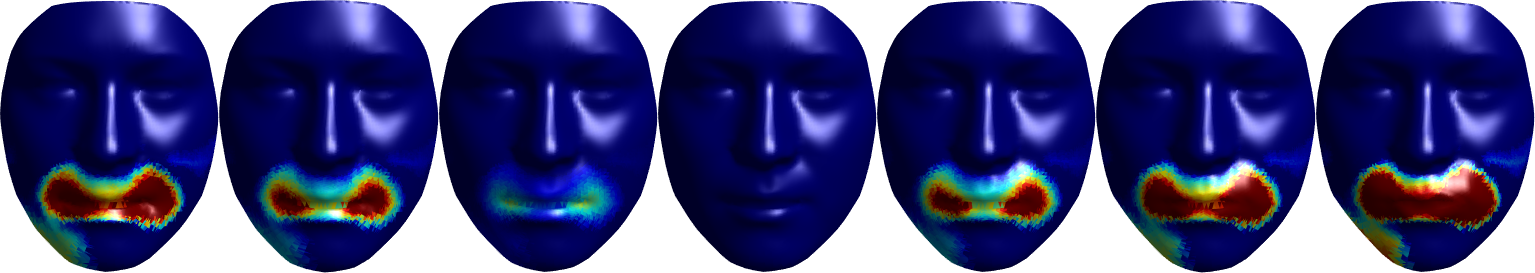}	
	\caption{\label{fig:3dmm_ex_slc} 3D shapes generated deforming the average model (middle column) by varying the first 4 SLC deformation components.}
\end{figure*}

In this section, we report additional examples of the new HLS-3DMM built from the $9,927$ registered scans of the FRGCv2.0, Bosphorus and FaceWarehouse datasets. We qualitatively analyze the latent space of the HLS-3DMM by applying some learned deformations to the average model $\mathbf{m}$. For a complete view, we construct both a PCA-based 3DMM and another one by learning our sparse SLC components. 3D faces generated deforming the average model by varying some PCA and SLC components are shown in Figure~\ref{fig:3dmm_ex_pca} and~\ref{fig:3dmm_ex_slc}, respectively. The accuracy of the proposed dense annotation transfer is evidenced in both the cases by the capability of the newly constructed 3DMM to reproduce complex expressions (Figures~\ref{fig:3dmm_ex_pca},~\ref{fig:3dmm_ex_slc} top row) as well as global shape variations as induced by different individuals and expression-like patterns. The smoothness of both the average model and the deformation components further demonstrates the effectiveness of the approach.

\begin{figure}[!t]
	\centering
	\includegraphics[width=0.85\linewidth]{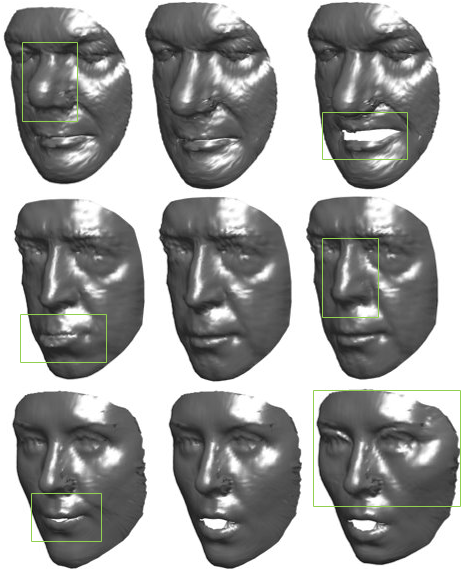}	
	\caption{\label{fig:def_transf} Qualitative examples of local deformations applied to registered scans (middle column) by means of our SLC components. Each deformation is obtained using a single component.}
\end{figure}

\subsection{Manipulating Registered Faces}\label{subsubsec:def_transf}
Another common way to qualitatively assess the fidelity of a dense registration algorithm consists in manipulating the registered 3D faces. To this aim, we learn the SLC components from the $9,927$ registered scans. Then, we deform some of the registered scans by applying randomly chosen SLC components (each deformation is obtained using a single component). In Figure~\ref{fig:def_transf}, few examples are shown. We can observe that local areas are smoothly warped with realistic and natural deformations. This suggests a semantic consistency was maintained during registration. In addition, the deformation results are accurate even if they are applied to non-neutral scans (Figure~\ref{fig:def_transf}, bottom row). Finally, we note that identity (nose shape, cheek size) as well as expression (mouth opening/closing) deformations can be performed.

\bibliographystyle{IEEEtran}
\bibliography{bibliography}
